\documentclass{article}

\PassOptionsToPackage{numbers}{natbib}
\usepackage[main, final]{neurips_2026}

\usepackage[utf8]{inputenc} 
\usepackage[T1]{fontenc}    
\usepackage{hyperref}       
\usepackage{url}            
\usepackage{booktabs}       
\usepackage{amsfonts}       
\usepackage{nicefrac}       
\usepackage{microtype}      
\usepackage{xcolor}         

\usepackage{colortbl}
\usepackage{todonotes}
\usepackage{graphicx}
\usepackage{multirow}
\usepackage{enumitem}
\usepackage{amsmath, amsthm, amssymb, amsfonts}
\usepackage{mathtools,mathrsfs}
\usepackage{microtype}      
\usepackage{booktabs}       
\usepackage{graphicx}
\usepackage[dvipsnames]{xcolor}
\usepackage{hyperref}       
\hypersetup{
    colorlinks=true,
    linkcolor=RoyalBlue,
    citecolor=Green,
    urlcolor=blue
}
\usepackage[capitalize,noabbrev]{cleveref} \usepackage[most]{tcolorbox} 
\usepackage{todonotes}
\usepackage{algorithm}
\usepackage{algorithmic}

\newcommand{\Works}{\mathcal{W}}
\newcommand{\ForgetReq}{\mathcal{W}_f}
\newcommand{\RetainReq}{\mathcal{W}_r}

\newcommand{\ForgetData}{D_f}
\newcommand{\RetainData}{D_r}

\newcommand{\Curator}{\mathcal{A}}
\newcommand{\Unlearner}{\mathcal{U}}

\newcommand{\Corpus}{\mathcal{C}}
\newcommand{\CorpusMid}{\mathcal{C}_{\mathrm{mid}}}
\newcommand{\CorpusPre}{\mathcal{C}_{\mathrm{pre}}}
\newcommand{\CorpusCC}{\mathcal{C}_{\mathrm{CC25}}}

\newcommand{\ForgetMetric}{F}
\newcommand{\RetainMetric}{R}

\newcommand{\benchmark}{\textsc{CleanSlate} }

\theoremstyle{definition}

\theoremstyle{remark}

\definecolor{cg100}{RGB}{199,233,192}  
\definecolor{cg75}{RGB}{216,240,211}
\definecolor{cg50}{RGB}{232,246,230}
\definecolor{cg25}{RGB}{243,250,242}
\definecolor{cneutral}{RGB}{255,255,255}
\definecolor{cr25}{RGB}{254,240,237}
\definecolor{cr50}{RGB}{253,222,214}
\definecolor{cr75}{RGB}{252,200,187}
\definecolor{cr100}{RGB}{251,180,163}  
\definecolor{cavg}{RGB}{237,237,237}   

\title{What to Forget in Unlearning?\\
Forget Set Curation for Language Models}

\author{%
  Animesh Jha$^*$ \\
  Stanford University\\
  \texttt{animjha@cs.stanford.edu} \\
  \And
  Arpandeep Khatua$^*$ \\
  Stanford University\\
  \texttt{akhatua@cs.stanford.edu} \\
  \AND
  Youssef Allouah$^\dagger$ \\
  Stanford University\\
  \texttt{yallouah@cs.stanford.edu} \\
  \And
  Sanmi Koyejo$^\dagger$ \\
  Stanford University\\
  \texttt{sanmi@cs.stanford.edu} \\
}

\begin{document}

\maketitle
{\renewcommand{\thefootnote}{}%
\footnotetext{$^*$Equal contribution. Alphabetical order.\quad
              $^\dagger$Equal advising. Alphabetical order.}%
\setcounter{footnote}{0}}

\begin{abstract}
Machine unlearning aims to efficiently remove data, knowledge, or behaviors from a trained model, but most methods and benchmarks are built around data deletion, where the examples to forget are known. In realistic language-model deployments, a requester may ask a model to stop reproducing a song or book without knowing which spans, documents, quotations, or near-duplicates in a trillion-token corpus support that behavior. Moving beyond this setting, we introduce \textsc{CleanSlate}, a benchmark for \emph{forget set curation}: mapping a high-level unlearning request to the data passed to an unlearning algorithm. \textsc{CleanSlate} evaluates verbatim suppression of songs and books using model-specific extractability, content-grounded QA, and general capability evaluations. We test natural curation strategies based on BM25 and exact-match retrieval, then an evaluation-aware curator that uses target reference windows directly as a stress test. We find that retrieval-derived forget sets behave very differently across models and unlearning algorithms and do not produce selective suppression. The evaluation-aware curator nearly eliminates requested continuations, yet also substantially suppresses non-requested ones and causes model-dependent capability loss. These results show that practical unlearning is not only an optimization problem once a forget set is given: the data chosen for forgetting determines both what can be unlearnt and what else is damaged.
 \end{abstract}





\section{Introduction}
\label{sec:intro}

Machine unlearning aims to remove targeted data or behaviors from a trained model without retraining from scratch~\citep{ginart2019making,yao2024large}. For language models, this goal is increasingly relevant in privacy, safety, and copyright settings, where model owners may be asked to suppress particular outputs or remove the influence of specific data. Yet most unlearning methods and benchmarks study the problem only after a crucial input has been supplied: the forget set, or the examples used to drive the update~\citep{maini2024tofu,shi2024muse,li2024wmdp,dorna2025openunlearning}. Real requests may not arrive in this form. A requester may ask a model to stop reproducing a song, book, or other protected work, while the unlearning algorithm requires concrete spans or documents to optimize against. This gap leaves a missing upstream problem: how should a suppression request be mapped to the data used for unlearning?

We study this missing step as \emph{forget set curation}: selecting the intervention data passed to an unlearning algorithm from a suppression request. We focus on verbatim output suppression for culturally embedded works such as songs and books. This is narrower than concept unlearning: suppressing a work should not require erasing its author, plot, genre, or cultural context. The desired behavior is selective: protected continuations should become difficult to elicit, while factual knowledge about the work and unrelated capabilities remain intact. We operationalize this target using probabilistic extraction methods that test whether a model assigns high probability to an exact suffix conditioned on its prefix~\citep{hayes2024measuring,cooper2025extracting}. Thus, after unlearning, target continuations should become un-extractable without collateral damage to the extractability of non-requested content or degradation in content-grounded QA and general capabilities.

\begin{figure*}[t]
    \centering
\includegraphics
[width=1\linewidth]{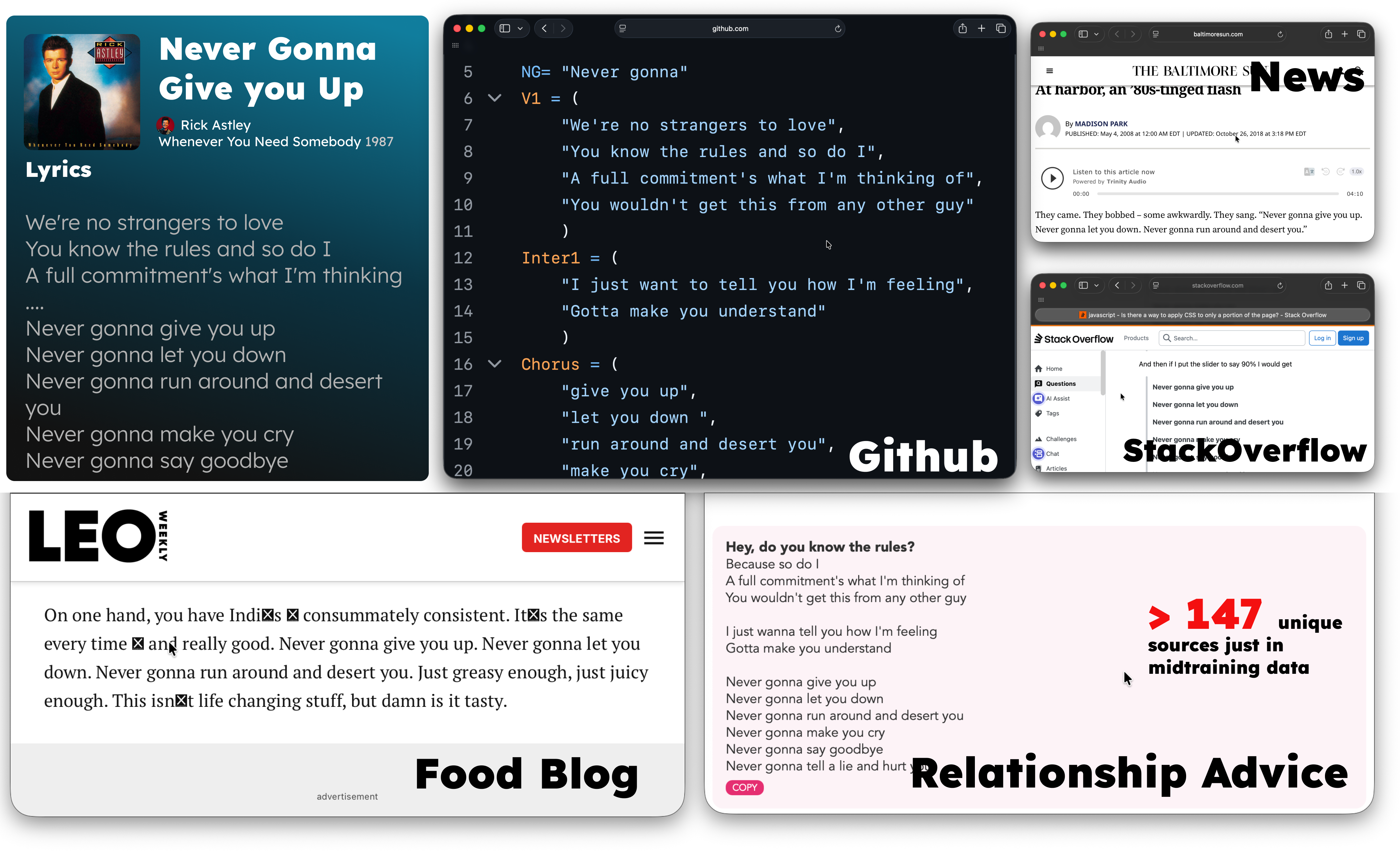}
\caption{\textit{Footprint of songs and books extends far beyond their canonical copy.} The top panel shows the canonical source for \textit{Never Gonna Give You Up}, with full lyrics. The surrounding panels show non-canonical sources that echo the same text, including a Python file on GitHub that stores the lyrics, a Baltimore Sun news article that weaves the hook into reporting on a flash mob, a Stack Overflow answer that uses the lines as filler text, a LEO Weekly food review that drops the chorus into prose about a sandwich, and a pickup lines site that recasts the lyrics as relationship advice.}
    \label{fig:cultural_embedding}
\end{figure*}

Songs and books expose why forget set curation is hard. Although a work may have a canonical text, the evidence supporting a model's continuation is rarely confined to that source. As shown in Figure~\ref{fig:cultural_embedding}, lyrics and passages can appear across training corpora through copies, quotations, reviews, fan forums, news articles, code snippets, synthetic examples, and incidental discussion. The relevant object is therefore the work's \emph{corpus footprint}: the distributed set of spans and documents that may support the target continuation. A curator must recover enough of this footprint to suppress the requested behavior, but not so much that it damages non-requested content or general capabilities. Exact matching helps reveal this footprint, but not fully since models can verbatim complete text even when exact n-gram matches have been removed from training data~\citep{liu2025languagemodelsverbatimcomplete}.

\textbf{Contributions.}
We conduct a study of \emph{forget set curation} for language-model unlearning: given a suppression request and a trained model, select the data that should be passed to an unlearning algorithm. This framing separates the request-to-data problem from the downstream model-update problem, and makes it possible to evaluate whether a proposed forget set actually induces selective suppression. To study this setting, we introduce \textsc{CleanSlate}, a benchmark and evaluation protocol for forget set curation. \textsc{CleanSlate} contains 4,616 Billboard Hot 100 songs from 1970--2025 and 50 books, with per-model extraction profiles for models spanning 7B--32B parameters. We add content-grounded QA pairs that distinguish suppression of verbatim reproduction from erasure of factual knowledge about the work. The benchmark evaluates the full pipeline: a curator receives a suppression request, returns intervention data, an unlearning algorithm updates the model, and the resulting model is evaluated for target suppression, collateral suppression, content-QA retention, and general capability retention. Our main findings are:
\begin{enumerate}[leftmargin=1.8em]
    \item \textbf{Cultural works have diffuse corpus footprints.} Songs and books are not represented only by canonical copies. Older works spread outward through quotes, reviews, code, fan forums, news, and incidental references; newer works also inherit older language through idioms, genre templates, public-domain quotations, and repeated phrases. Thus, the evidence that can support a target continuation may predate the work itself or appear in sources that do not look like copies.
    \item \textbf{Curator and unlearner interact strongly.} The same retrieval-derived forget set can produce near-zero or near-complete target suppression depending on the unlearner and model; across the tested grid, strong suppression is always accompanied by substantial collateral suppression.
    \item \textbf{Evaluation aware curation reveals a selectivity gap.} When we bypass retrieval and select the windows used for evaluation directly, unlearning can suppress requested continuations almost completely. However, it also causes substantial model-dependent capability regressions and suppresses non-requested content. Thus, even in verbatim output suppression, identifying the target text is not the same as constructing a clean forget set.
\end{enumerate}
Together, these results suggest that the forget set should be treated not as a premise of language-model unlearning, but as one of its objects of evaluation. What is selected for forgetting determines not only whether the requested behavior is suppressed, but also which neighboring abilities and general capabilities are disturbed. We argue that practical unlearning should therefore be studied as a pipeline from request, to curated forget set, to edited model.

\section{Related Work}
\label{sec:related}


\paragraph{Memorization and verbatim extraction.}
Language model memorization is commonly studied through extraction: prompting a model with a prefix and testing whether it reproduces a target suffix~\citep{carlini2021extracting,carlini2023quantifying,ahmed2026extracting}. \cite{hayes2024measuring} refine this into probabilistic discoverable extraction, measuring whether a target continuation can be produced under repeated sampling. \cite{cooper2025extracting} apply this framework to copyrighted books, showing that extractability varies substantially across works and model families. We use this completion-style notion of memorization as a proxy for verbatim output suppression, the goal is not to infer training membership or erase all knowledge of a work, but to make requested continuations difficult to elicit while preserving nearby knowledge and unrelated capabilities.

\paragraph{Machine unlearning benchmarks.}
Machine unlearning aims to remove the influence of specified data or behaviors from a trained model without retraining from scratch~\citep{ginart2019making,yao2024large}. A range of methods have been proposed for language models, including loss ascent, preference-based objectives, logit adjustment, and representation-level interventions~\citep{zhang2024negative,li2024wmdp,fan2025simplicityprevailsrethinkingnegative,dong2025undial}. Existing benchmarks evaluate whether such updates suppress targeted information while preserving utility, including synthetic biographies in TOFU~\citep{maini2024tofu}, multi-axis evaluation in MUSE~\citep{shi2024muse}, hazardous-knowledge unlearning in WMDP~\citep{li2024wmdp}, and unified evaluation in OpenUnlearning~\citep{dorna2025openunlearning}. These benchmarks differ in domain and objective, but they typically provide the examples, entities, or evaluation targets to be forgotten. In contrast, we study the upstream curation problem: given a suppression request for a work, what data should be selected for the unlearning update?

\paragraph{Data selection, retrieval, and attribution.}
Recent work shows that the contents of a forget set matter even after the forget data has been specified: small subsets or token-level selections can substantially change the suppression--preservation tradeoff~\citep{pal2025llm,wan2025not,zhou2025tokensmeantforgotten,allouah2026distributional,liu2026randomized}. This motivates studying selection itself, but prior work largely assumes that the unwanted examples or target domain are already known. A natural approach to request-level curation is to retrieve text overlapping with the target work, using lexical search or exact-substring systems such as BM25, Infini-gram, and Infini-gram-mini~\citep{liu2024infini,xu2025infini}. However, textual overlap is only an imperfect proxy for the data responsible for a model continuation: models can verbatim complete text even when exact n-gram matches have been removed from training data~\citep{liu2025languagemodelsverbatimcomplete}. Influence functions and datamodeling offer a more causal view of training-example responsibility~\citep{koh2017understanding,ilyas2022datamodels,engstrom2024dsdm,georgiev2024attribute}, but applying them to request-level curation over trillion-token corpora remains an open challenge.

\section{Problem Statement: Forget Set Curation}
\label{sec:setup}
Let $\theta$ be a pretrained language model, and let $\Works$ be a collection of works, such as songs or books. A suppression request identifies a target subset $\ForgetReq \subset \Works$ whose verbatim reproduction should be suppressed. Any possible verbatim reproduction of works in $\RetainReq = \Works \setminus \ForgetReq$ should be preserved.
\paragraph{Extractability.} We quantify verbatim reproduction through probabilistic extraction \citep{hayes2024measuring, cooper2025extracting}. A work is divided into a sequence of prefix-suffix window pairs $(x,z)$. For a given prefix $x$, the model assigns a probability 
$$p_z \;=\; p_\theta(z \mid x) \;=\; \prod_{t=1}^{|z|} p_\theta(z_t \mid x,\, z_{<t})$$ 
We call a window as \emph{extractable} if $p_z \geq \tau$, we use $\tau = 0.001$ from~\citep{cooper2025extracting}. 
\paragraph{Curated forget sets.}
Given a suppression request $\ForgetReq$, target texts corresponding to $\ForgetReq$, a trained model $\theta$, and access to a large search corpus $\Corpus$, a curator $\Curator$ returns a forget set $\ForgetData$ and, optionally, a retain set $\RetainData$. A downstream unlearning algorithm $\Unlearner$ then produces an unlearnt model 
\[\theta' = \Unlearner(\theta, \ForgetData, \RetainData).\]
In this paper, the curator $\Curator$ is the object under evaluation: we compare different choices of $\Curator$ while holding the downstream unlearning procedure fixed unless otherwise stated.
\paragraph{Search Corpora.} For our analysis and experiments we use three distinct scale corpora, $\CorpusMid$ the \texttt{Dolmino} midtraining mix~\citep{olmo2025olmo}, $\CorpusPre$ a subset of the \texttt{Dolma3} pretraining mix~\citep{olmo2025olmo}, and $\CorpusCC$ the Jan 2025 Common Crawl snapshot. See Appendix~\ref{appendix:corpora_details} for details.
\paragraph{Evaluation.}
The goal of verbatim output suppression is to make extractable windows from $\ForgetReq$ un-extractable after unlearning, without inducing the same effect on extractable windows from $\RetainReq$. Verbatim suppression should not erase knowledge about a work, we also evaluate content-grounded question answering over the same works, together with general capability benchmarks. Thus, a curator is judged not by textual relevance alone, but by the behavior of the unlearnt model it induces. \Cref{fig:cleanslate_overview} captures the \benchmark pipeline.
The central difficulty is that the documents in $\Corpus$ that support a target continuation need not be canonical copies of the requested work. We next show that songs and books often have broad corpus footprints, making the mapping from $\ForgetReq$ to $\ForgetData$ nontrivial.
\begin{figure*}[t]
    \centering
\includegraphics[width=\linewidth]{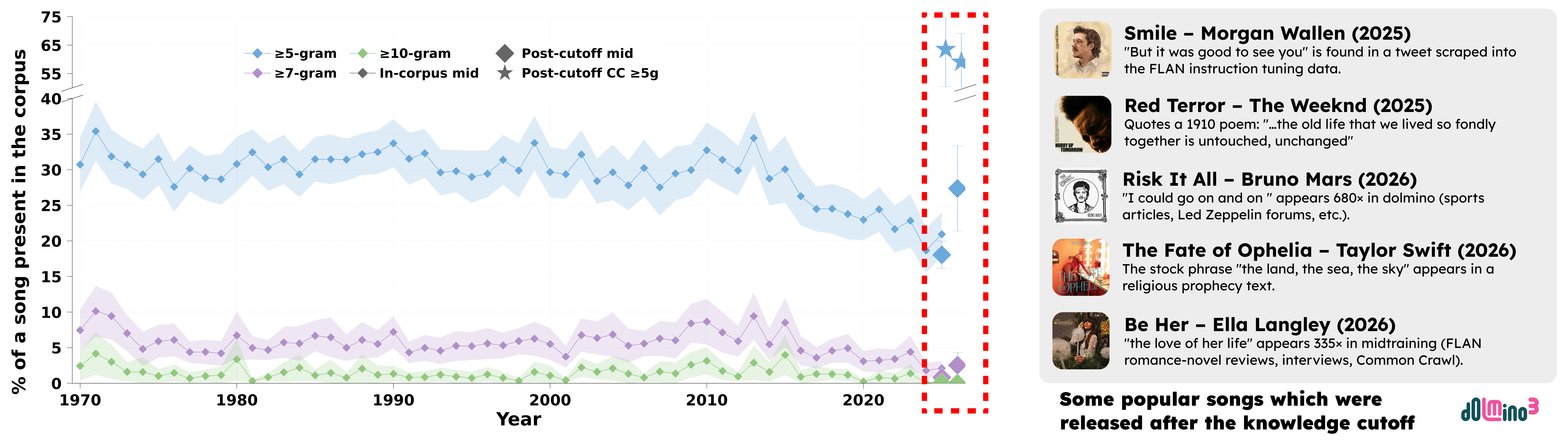}
\caption{\textit{Corpus footprints arise through both outward and inward diffusion.} \textbf{(Left)} Median per-song coverage in midtraining corpus across their release years, 1970 to 2026. Older works often have broad footprints from copies, quotations, discussion, and other web sources. These are cases of outward diffusion. \textbf{(Right)} Post-cutoff markers show coverage for songs released after cutoff of the corpus. We present five examples, each match older or unrelated sources, including instruction-tuning data, a 1910 poem, forums, religious text, and reviews. These cases illustrate inward diffusion: new works can inherit phrases, quotations, or stock language already present in the corpus.
}
    \label{fig:year_mem}
\end{figure*}

\section{The Corpus Footprint: Why Forget Set Curation Is Hard} 
\label{sec:cultural_embedding}
The training evidence supporting a verbatim continuation need not be isolated to a canonical copy of the work. It may appear in lyric aggregators, forum discussions, fan fiction, or code snippets. A curator's true target is therefore a work's \emph{corpus footprint}: the distributed set of documents and spans that can support the target continuation.
\paragraph{Measuring literal overlap.}
To quantify the corpus footprint of a work, we measure the scale of exact word-level overlaps between the target text and a corpus $\Corpus$ using Infini-gram-mini~\citep{xu2025infini}. For a given work, we compute localized $n$-gram occurrence counts across every position (see Appendix~\ref{appendix:coverage_comp} for details). This is a conservative measure as changes in formatting, whitespace, or even case can break a match. It also misses paraphrases, translations, and semantic references, nor does it identify which documents actually caused a continuation. Instead, it lower bounds the literal overlap that a curator would miss if it searched only for canonical copies. At web scale, any large corpus will contain many short n-grams from almost any English text even when no canonical copy is present. So in addition we also measure  the \emph{coverage}: the fraction of the requested work covered by $N$-gram matches. This tracking reveals two primary mechanisms of data diffusion, unpacked below.

\paragraph{Outward diffusion.} Works with cultural impact quickly spread after release. \texttt{Smile} (Morgan Wallen, released 31st Dec 2024) illustrates how rapidly a footprint forms: in $\CorpusCC$ (Jan 2025), it exhibits $100\%$ coverage at the 5-gram level, with some spans having $\geq 10^5$ occurrences, it also has $100\%$ coverage at the 50-gram level with some spans having  $\geq 10^4$  matches (see Figure~\ref{fig:coverage_smile}). Clearly $\CorpusCC$ contains canonical copies of \texttt{Smile}, but also a large cloud of shorter exact overlaps. Older works have had decades to diffuse, they are copied, quoted, remixed, discussed, and embedded in unexpected sources. Figure~\ref{fig:cultural_embedding} illustrates this for \textit{Never Gonna Give You Up}: beyond canonical lyric pages, the same text appears in fan transcriptions, wiki pages, code files, review prose, and question-answering sites. In $\CorpusMid$, this work has $94.2\%$ coverage at $n=5$ and $56.4\%$ coverage at $n=25$ (see Figure~\ref{fig:mem_coverage_song_nggyu}), with matched spans distributed across $\geq147$ unique documents. Any curator looking solely at canonical sources would miss the vast majority of these spans.


\paragraph{Inward diffusion.} A new work may contain language that already existed in the corpus before the work itself was released. This can happen through idioms, genre templates, stock phrases, public-domain quotations, or repeated cultural language. Figure~\ref{fig:year_mem} shows several post-cutoff songs with substantial exact-overlap coverage in corpora that predate their release. For example, \textit{Red Terror} by The Weeknd has roughly 20\% coverage at $n=7$ in $\CorpusMid$ because it quotes a 1910 poem. This shows that parts of the requested continuation may be supported by older or unrelated text. This makes curation more subtle than finding noisy and approximate copies of the target work: the relevant evidence may predate the work or appear in documents that do not look like copies at all.

\paragraph{Connection to target continuations.}
Corpus footprints help explain where verbatim output suppression becomes difficult. Figures~\ref{fig:mem_coverage_song_nggyu}--\ref{fig:mem_coverage_communist_manifesto} overlay localized corpus occurrence counts with extraction probabilities across songs, poems, and books. Across these case studies, extractable continuations often occur in regions with dense literal overlap, such as song choruses, famous quotations, or repeated phrases. We treat this as a diagnostic rather than a causal claim: high overlap does not prove that a particular document caused a completion, and low overlap does not rule out elicitation through other mechanisms. The practical implication is that a curator that misses dense regions of the footprint may leave the model enough data to be able to reproduce the requested continuation, while a curator that captures them too broadly may also affect neighboring content.

\section{\benchmark} \label{sec:benchmark}

\begin{figure*}[t]
    \centering
    \includegraphics[width=\linewidth]{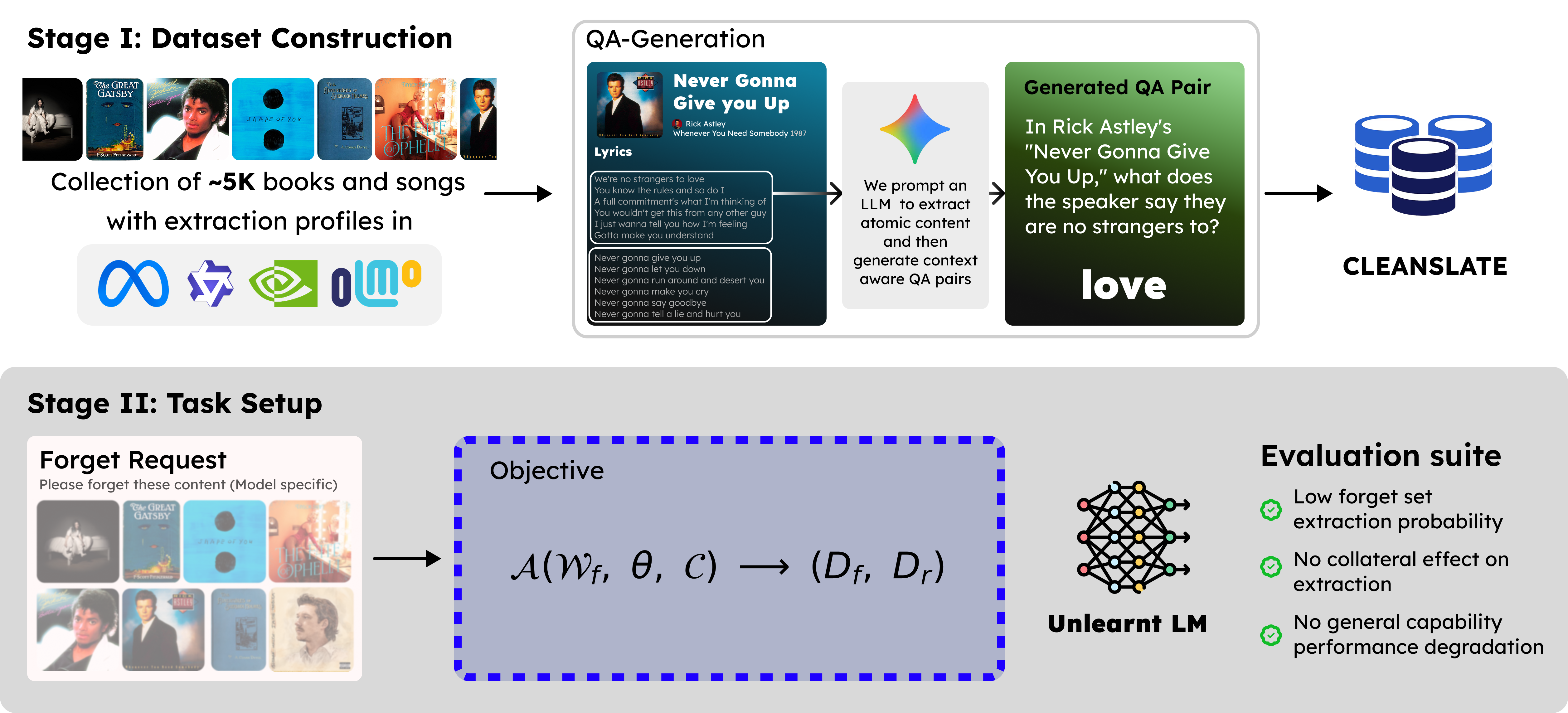}
\caption{\textbf{CleanSlate overview.} \textit{(Top)} We assemble $\sim$5K books and songs and measure baseline extraction profiles across different model families and generate QA pairs for each content. Together these form the CleanSlate dataset. \textit{(Bottom)} Given a forget request $\mathcal{W}_f$ specifying $k$ items, target model parameters $\theta$, and an arbitrary corpus $\mathcal{C}$ (real, synthetic, or generated), the objective is to create a curator $\mathcal{A}$ that outputs a forget set $\mathcal{D}_f$ and retain set $\mathcal{D}_r$. These are passed to an unlearning procedure, and the resulting model is evaluated for low forget set extraction, preserved extraction on retained content, and stable general capability performance.}
    \label{fig:cleanslate_overview}
\end{figure*}

Building on the formalism of Section~\ref{sec:setup}, \benchmark pairs each suppression request with per-model extractability evidence and content-grounded QA, so that the full pipeline of curator, unlearning algorithm, and edited model can be scored along several axes. Figure~\ref{fig:cleanslate_overview} summarizes the construction and evaluation flow.

\paragraph{Content domains.}
\benchmark covers two domains, songs and books, whose corpus footprints (Section~\ref{sec:cultural_embedding}) are shaped differently. Songs travel as short, repetitive, discrete units. Books propagate as longer-form passages reaching the corpus primarily through commentary and excerption rather than full copies. A curator that recovers one shape may miss most of the other, making the two domains complementary stress tests. We source songs from Billboard Hot 100 annual charts spanning 1970--2025, matched against the LRCLib lyrics database by title and artist, and 50 books, mostly from Project Gutenberg~\citep{gutenberg} with a small number of closed-license works.

\paragraph{Extraction profiles.}
We slide a window of 100 prefix and 100 suffix characters with stride 10 over each work, and label a work \emph{model-extractable} for a base model $\theta$ if at least $5\%$ of its windows are extractable in the sense of Section~\ref{sec:setup}. The resulting profile is per-model. The same song may be extractable for one model and not for another. See Appendix~\ref{appendix:memorization_flame} for more details.

\paragraph{Forget and retain pools.}
A suppression request is a model specific sample of size $|\ForgetReq|=50$ drawn from the model-extractable works of $\theta$, with $\RetainReq$ defined as in Section~\ref{sec:setup}. For a fixed model we reuse the same $\ForgetReq$ and $\RetainReq$ across curators so that differences are attributable to the curation.

\paragraph{Content-grounded QA.} \label{sec:qa}
Suppressing verbatim reproduction and erasing factual knowledge are distinct goals. A model that has been asked to stop reproducing a song's lyrics should still be able to answer factual questions about its content. We construct CleanSlate-QA in two stages. First, we prompt an LLM\footnote{\texttt{gemini-3.1-flash-lite-preview} via LiteLLM; Appendix~\ref{app:qa_construction} lists all models used.} to extract \emph{atomic statements} from each work; factual statements anchored to a named entity, place, number, or concrete event mentioned inside the text, with priors such as title, creator, year, and genre explicitly excluded. Second, each statement is turned into a QA pair whose question embeds the title and creator naturally so it is self-contained, and whose answer is a short 1-5 word entity. For example, in J.K. Rowling's \emph{Harry Potter and the Sorcerer's Stone} we ask where the Dursleys make Harry sleep, with answer \emph{cupboard under the stairs}. The final dataset contains 12{,}088 QA pairs spanning the songs and books in $\Works$. 

\paragraph{End-to-end evaluation.}
\label{sec:eval_protocol}
For each curator $\Curator$, we measure baseline extractability and QA on $\theta$, run $\Curator$ to obtain $\ForgetData$, apply an unlearning algorithm $\Unlearner$ to obtain $\theta'$, and re-measure. $\ForgetMetric\uparrow$ is the net percentage reduction in the number of extractable windows of works in $\ForgetReq$ from $\theta$ to $\theta'$; $\RetainMetric\downarrow$ is the same quantity over $\RetainReq$, where high values indicate collateral suppression. Negative values denote a net increase. \textit{QA $\Delta$} is the change in CleanSlate-QA accuracy. A validation suite covers math (GSM8K~\citep{cobbe2021trainingverifierssolvemath}), held-out reasoning (BBH~\citep{suzgun2022challengingbigbenchtaskschainofthought}), commonsense (WinoGrande~\citep{sakaguchi2019winograndeadversarialwinogradschema}), reading comprehension (CoQA~\citep{reddy2019coqaconversationalquestionanswering}), code (HumanEval+~\citep{liu2023codegeneratedchatgptreally}), and language modeling (LAMBADA~\citep{radford2019language}). Per-model baselines are reported in Appendix~\ref{appendix:memorization_flame}. \benchmark supports two evaluation modes: request-level evaluation lets curator and unlearner vary jointly and scores the composed request-to-data-to-update pipeline; algorithm-focused evaluation holds two of $\langle$curator, unlearner, model$\rangle$ fixed and varies the third.

\section{Experiments}
\label{sec:experiments}
We evaluate three retrieval curators and an evaluation-aware stress test using \benchmark with $|\ForgetReq|=50$. Table~\ref{tab:retrieval} fixes the unlearner at SimNPO so row differences isolate the curator. Table~\ref{tab:algos} fixes $\Curator$ to the evaluation aware curator and varies the unlearner $\Unlearner$, (see Appendix~\ref{sec:unlearning_hparams} for hyperparameter details). Table~\ref{tab:cross} completes the curator $\times$ unlearner grid on three models. We perform experiments over six models, \textsc{Llama-3.1-8B}, \textsc{Olmo-3-7B},  \textsc{Nemotron-9B}, \textsc{Qwen3-8B},  \textsc{Gemma-3-12B}, and \textsc{Olmo-3-32B}. Per-model pre-unlearning evaluations are in Appendix Table~\ref{tab:baseline}.

\begin{table}[t]
\centering
\footnotesize
\setlength{\tabcolsep}{4pt}
\renewcommand{\arraystretch}{1.1}
\caption{Curation comparison at $|\ForgetReq|=50$ with SimNPO unlearning, for three retrieval-based curators (BM25 pre, BM25 mid, Infini-gram mid) and the evaluation-aware curator (EA). $F$ and $R$ are net percentage reductions in extractable-window counts on the forget and retain pools; negative values denote net increases (Section~\ref{sec:eval_protocol}). QA and validation columns report percentage-point changes from the corresponding model baselines (Table~\ref{tab:baseline}); \textit{Avg} averages the six validation changes.}
\label{tab:retrieval}
\begin{tabular}{ll|cc|c|cccccc|c}
\toprule
& & \multicolumn{2}{c|}{Net reduction (\%)} & QA & \multicolumn{7}{c}{Validation $\Delta$ (pp)} \\
\cmidrule(lr){3-4} \cmidrule(lr){6-12}
& Model & F\,$\uparrow$ & R\,$\downarrow$ & $\Delta$ (pp) & GSM8K & BBH & WinoG & CoQA & HEval+ & LMBDA & \textit{Avg} \\
\midrule
\multirow{7}{*}{\rotatebox[origin=c]{90}{BM25 pre}}
& \textsc{Gemma-3-12B}    & \cellcolor{cg50} 39.3\% & \cellcolor{cr50} 38.4\% & \cellcolor{cr50} -2.3 & \cellcolor{cg75} +3.9 & \cellcolor{cr100} -4.7 & \cellcolor{cg25} +0.2 & \cellcolor{cr100} -6.4 & \cellcolor{cg100} +4.9 & \cellcolor{cr100} -4.9 & \cellcolor{cr25} -1.2 \\
& \textsc{Olmo-3-32B}     & \cellcolor{cg50} 43.1\% & \cellcolor{cr50} 30.8\% & \cellcolor{cr25} -0.2 & \cellcolor{cg50} +1.5 & \cellcolor{cg25} +0.2 & \cellcolor{cg25} +0.6 & \cellcolor{cg100} +4.6 & \cellcolor{cneutral} +0.0 & \cellcolor{cg25} +0.5 & \cellcolor{cg25} +1.2 \\
& \textsc{Llama-3.1-8B}   & \cellcolor{cg50} 28.9\% & \cellcolor{cr75} 51.2\% & \cellcolor{cr25} -0.4 & \cellcolor{cr25} -0.1 & \cellcolor{cg25} +0.1 & \cellcolor{cr25} -0.6 & \cellcolor{cg50} +1.9 & \cellcolor{cg25} +1.2 & \cellcolor{cr25} -1.0 & \cellcolor{cg25} +0.3 \\
& \textsc{Nemotron-9B}    & \cellcolor{cg25} 14.6\% & \cellcolor{cr25} 14.9\% & \cellcolor{cg25} +0.6 & \cellcolor{cr100} -11.1 & \cellcolor{cr75} -4.0 & \cellcolor{cg25} +0.9 & \cellcolor{cg25} +0.6 & \cellcolor{cg25} +0.6 & \cellcolor{cr50} -1.7 & \cellcolor{cr50} -2.4 \\
& \textsc{Qwen3-8B}       & \cellcolor{cg25} 1.6\% & \cellcolor{cr25} 2.1\% & \cellcolor{cr75} -3.9 & \cellcolor{cr25} -1.4 & \cellcolor{cg25} +1.4 & \cellcolor{cg75} +3.2 & \cellcolor{cr75} -3.5 & \cellcolor{cr100} -6.7 & \cellcolor{cg75} +3.8 & \cellcolor{cr25} -0.5 \\
& \textsc{Olmo-3-7B}      & \cellcolor{cg25} 1.0\% & \cellcolor{cr25} 2.9\% & \cellcolor{cr25} -0.1 & \cellcolor{cg50} +1.7 & \cellcolor{cr25} -0.1 & \cellcolor{cr25} -0.1 & \cellcolor{cr75} -3.8 & \cellcolor{cr25} -0.6 & \cellcolor{cneutral} -0.0 & \cellcolor{cr25} -0.5 \\
& \textit{Average}     & \cellcolor{cavg} 21.4\% & \cellcolor{cavg} 23.4\% & \cellcolor{cavg} -1.1 & \cellcolor{cavg} -0.9 & \cellcolor{cavg} -1.2 & \cellcolor{cavg} +0.7 & \cellcolor{cavg} -1.1 & \cellcolor{cavg} -0.1 & \cellcolor{cavg} -0.5 & \cellcolor{cavg} -0.5 \\
\midrule
\midrule
\multirow{7}{*}{\rotatebox[origin=c]{90}{BM25 mid}}
& \textsc{Gemma-3-12B}    & \cellcolor{cg50} 46.3\% & \cellcolor{cr50} 46.0\% & \cellcolor{cr75} -3.2 & \cellcolor{cg50} +2.0 & \cellcolor{cr100} -8.6 & \cellcolor{cr25} -0.4 & \cellcolor{cr100} -9.2 & \cellcolor{cg50} +2.4 & \cellcolor{cr100} -6.3 & \cellcolor{cr75} -3.4 \\
& \textsc{Olmo-3-32B}     & \cellcolor{cg50} 43.0\% & \cellcolor{cr50} 35.2\% & \cellcolor{cg25} +0.3 & \cellcolor{cg50} +2.4 & \cellcolor{cg25} +0.3 & \cellcolor{cg25} +1.4 & \cellcolor{cg50} +2.9 & \cellcolor{cg25} +0.6 & \cellcolor{cg25} +0.7 & \cellcolor{cg25} +1.4 \\
& \textsc{Llama-3.1-8B}   & \cellcolor{cg25} 16.7\% & \cellcolor{cr75} 54.0\% & \cellcolor{cr25} -1.1 & \cellcolor{cr25} -0.2 & \cellcolor{cr25} -0.4 & \cellcolor{cr25} -1.5 & \cellcolor{cr50} -2.4 & \cellcolor{cg50} +1.8 & \cellcolor{cr25} -1.4 & \cellcolor{cr25} -0.7 \\
& \textsc{Nemotron-9B}    & \cellcolor{cg25} 13.0\% & \cellcolor{cr25} 14.8\% & \cellcolor{cg25} +0.6 & \cellcolor{cr100} -10.1 & \cellcolor{cr75} -4.1 & \cellcolor{cg25} +0.6 & \cellcolor{cr25} -0.3 & \cellcolor{cg25} +0.6 & \cellcolor{cr50} -2.4 & \cellcolor{cr50} -2.6 \\
& \textsc{Qwen3-8B}       & \cellcolor{cg25} 2.3\% & \cellcolor{cr25} 2.7\% & \cellcolor{cr100} -5.8 & \cellcolor{cr25} -1.1 & \cellcolor{cg25} +1.4 & \cellcolor{cg75} +3.6 & \cellcolor{cr25} -0.6 & \cellcolor{cr100} -10.4 & \cellcolor{cg75} +3.6 & \cellcolor{cr25} -0.6 \\
& \textsc{Olmo-3-7B}      & \cellcolor{cg25} 1.5\% & \cellcolor{cr25} 3.7\% & \cellcolor{cneutral} -0.0 & \cellcolor{cg25} +1.0 & \cellcolor{cr25} -0.2 & \cellcolor{cneutral} +0.0 & \cellcolor{cr50} -1.9 & \cellcolor{cr25} -1.2 & \cellcolor{cr25} -0.3 & \cellcolor{cr25} -0.4 \\
& \textit{Average}     & \cellcolor{cavg} 20.5\% & \cellcolor{cavg} 26.1\% & \cellcolor{cavg} -1.5 & \cellcolor{cavg} -1.0 & \cellcolor{cavg} -2.0 & \cellcolor{cavg} +0.6 & \cellcolor{cavg} -1.9 & \cellcolor{cavg} -1.0 & \cellcolor{cavg} -1.0 & \cellcolor{cavg} -1.0 \\
\midrule
\multirow{7}{*}{\rotatebox[origin=c]{90}{Infinigram mid}}
& \textsc{Gemma-3-12B}    & \cellcolor{cg50} 47.8\% & \cellcolor{cr50} 44.9\% & \cellcolor{cr50} -2.8 & \cellcolor{cg75} +3.2 & \cellcolor{cr100} -5.3 & \cellcolor{cr25} -0.2 & \cellcolor{cr100} -10.1 & \cellcolor{cg50} +1.8 & \cellcolor{cr100} -6.0 & \cellcolor{cr50} -2.8 \\
& \textsc{Olmo-3-32B}     & \cellcolor{cg50} 44.9\% & \cellcolor{cr50} 36.0\% & \cellcolor{cg25} +0.4 & \cellcolor{cg50} +2.7 & \cellcolor{cg25} +0.1 & \cellcolor{cg50} +1.7 & \cellcolor{cg75} +3.9 & \cellcolor{cneutral} +0.0 & \cellcolor{cg25} +0.9 & \cellcolor{cg50} +1.5 \\
& \textsc{Llama-3.1-8B}   & \cellcolor{cg25} 19.4\% & \cellcolor{cr75} 55.7\% & \cellcolor{cr25} -0.8 & \cellcolor{cg25} +0.3 & \cellcolor{cr25} -0.4 & \cellcolor{cr25} -0.9 & \cellcolor{cg25} +0.9 & \cellcolor{cneutral} +0.0 & \cellcolor{cr25} -1.4 & \cellcolor{cr25} -0.2 \\
& \textsc{Nemotron-9B}    & \cellcolor{cg25} 12.7\% & \cellcolor{cr25} 14.5\% & \cellcolor{cg25} +0.5 & \cellcolor{cr25} -0.1 & \cellcolor{cr50} -2.3 & \cellcolor{cg25} +0.2 & \cellcolor{cg25} +0.3 & \cellcolor{cg75} +3.7 & \cellcolor{cr50} -2.3 & \cellcolor{cr25} -0.1 \\
& \textsc{Qwen3-8B}       & \cellcolor{cg25} 2.8\% & \cellcolor{cr25} 2.8\% & \cellcolor{cr100} -4.6 & \cellcolor{cr25} -1.0 & \cellcolor{cg25} +1.5 & \cellcolor{cg75} +3.9 & \cellcolor{cr75} -3.1 & \cellcolor{cr100} -9.8 & \cellcolor{cg75} +3.6 & \cellcolor{cr25} -0.8 \\
& \textsc{Olmo-3-7B}      & \cellcolor{cg25} 1.6\% & \cellcolor{cr25} 3.2\% & \cellcolor{cr25} -0.1 & \cellcolor{cg25} +0.9 & \cellcolor{cr25} -0.3 & \cellcolor{cneutral} +0.0 & \cellcolor{cr25} -0.8 & \cellcolor{cr25} -0.6 & \cellcolor{cr25} -0.1 & \cellcolor{cr25} -0.2 \\
& \textit{Average}     & \cellcolor{cavg} 21.6\% & \cellcolor{cavg} 26.2\% & \cellcolor{cavg} -1.2 & \cellcolor{cavg} +1.0 & \cellcolor{cavg} -1.1 & \cellcolor{cavg} +0.8 & \cellcolor{cavg} -1.5 & \cellcolor{cavg} -0.8 & \cellcolor{cavg} -0.9 & \cellcolor{cavg} -0.4 \\
\midrule
\midrule
\multirow{7}{*}{\rotatebox[origin=c]{90}{EA}}
& \textsc{Gemma-3-12B}    & \cellcolor{cg100} 96.2\% & \cellcolor{cr100} 85.4\% & \cellcolor{cr100} -6.1 & \cellcolor{cg75} +3.6 & \cellcolor{cr100} -29.4 & \cellcolor{cneutral} +0.0 & \cellcolor{cr100} -6.3 & \cellcolor{cg25} +1.2 & \cellcolor{cr100} -13.8 & \cellcolor{cr100} -7.4 \\
& \textsc{Olmo-3-32B}     & \cellcolor{cg100} 100.0\% & \cellcolor{cr100} 97.0\% & \cellcolor{cr50} -2.7 & \cellcolor{cg75} +3.2 & \cellcolor{cr75} -3.3 & \cellcolor{cr100} -4.9 & \cellcolor{cg75} +3.9 & \cellcolor{cr25} -1.2 & \cellcolor{cr75} -3.0 & \cellcolor{cr25} -0.9 \\
& \textsc{Llama-3.1-8B}   & \cellcolor{cg100} 100.0\% & \cellcolor{cr100} 100.0\% & \cellcolor{cr100} -5.7 & \cellcolor{cr100} -11.8 & \cellcolor{cr100} -37.6 & \cellcolor{cr75} -3.1 & \cellcolor{cr75} -3.9 & \cellcolor{cg75} +3.7 & \cellcolor{cr100} -10.1 & \cellcolor{cr100} -10.4 \\
& \textsc{Nemotron-9B}    & \cellcolor{cg100} 100.0\% & \cellcolor{cr100} 80.3\% & \cellcolor{cr25} -1.2 & \cellcolor{cr100} -23.1 & \cellcolor{cr100} -22.8 & \cellcolor{cg25} +0.2 & \cellcolor{cr100} -5.5 & \cellcolor{cr25} -1.2 & \cellcolor{cr100} -5.2 & \cellcolor{cr100} -9.6 \\
& \textsc{Qwen3-8B}       & \cellcolor{cg100} 100.0\% & \cellcolor{cr100} 82.7\% & \cellcolor{cr100} -21.9 & \cellcolor{cr50} -3.0 & \cellcolor{cr50} -2.0 & \cellcolor{cg25} +0.3 & \cellcolor{cr100} -7.9 & \cellcolor{cr100} -4.9 & \cellcolor{cr100} -6.0 & \cellcolor{cr75} -3.9 \\
& \textsc{Olmo-3-7B}      & \cellcolor{cg100} 100.0\% & \cellcolor{cr75} 72.2\% & \cellcolor{cr25} -1.0 & \cellcolor{cg50} +2.8 & \cellcolor{cr25} -0.1 & \cellcolor{cr50} -2.0 & \cellcolor{cr100} -5.4 & \cellcolor{cg50} +1.8 & \cellcolor{cr50} -2.3 & \cellcolor{cr25} -0.9 \\
& \textit{Average}     & \cellcolor{cavg} 99.4\% & \cellcolor{cavg} 86.3\% & \cellcolor{cavg} -6.5 & \cellcolor{cavg} -4.7 & \cellcolor{cavg} -15.9 & \cellcolor{cavg} -1.6 & \cellcolor{cavg} -4.2 & \cellcolor{cavg} -0.1 & \cellcolor{cavg} -6.7 & \cellcolor{cavg} -5.5 \\
\bottomrule
\end{tabular}
\end{table}

\paragraph{Curators.}
To evaluate whether standard corpus search can construct selective forget sets, we test two retrieval-based curators against large training corpora, alongside an evaluation-aware baseline:   \textbf{BM25-pre} and \textbf{BM25-mid} use BM25 indices over chunked documents from $\CorpusPre$ and $\CorpusMid$, respectively. For each requested work, the curator keeps the top ranked retrieval units and converts them into completion examples. \textbf{Infini-gram-mid} uses an Infini-gram-mini index over $\CorpusMid$ to find maximal exact character spans of the requested text that occur in the corpus. The matched spans are then localized in retrieved corpus contexts. Finally, we test an Evaluation Aware \textbf{(EA)} curator which for each work in $\ForgetReq$, constructs $\ForgetData$ to contain all the windows for that reference-text. EA presumes access to the evaluation windows and serves as a robustness test of the evaluation rather than a practical curator. All curators generate $\ForgetData$ in the same format: a 100-character prefix followed by a 100-character suffix. For BM25, we slide this window over the retrieved BM25 text units recorded in the retrieval artifact. For Infini-gram, we emit windows from the retrieved corpus context whose 200-character frame contains the midpoint of the exact match. This fixes sequence length but not the number of examples: the natural curator outputs are not size matched. We treat selected content and intervention volume as properties of the end-to-end curator output; with fixed epochs, output size also changes the number of optimizer updates. Exact output sizes are reported in Appendix~\ref{app:df_sizes}, and further retrieval details are in Appendix~\ref{app:retrieval_details}.
\subsection{Can natural retrieval curators induce selective suppression?}
\label{sec:exp-retrieval}

Under SimNPO, we find that retrieval engagement broadly tracks model size (Table~\ref{tab:retrieval}). The two largest models (\textsc{Gemma-3-12B}, \textsc{Olmo-3-32B}) show the strongest forget-side movement, \textsc{Llama-3.1-8B} and \textsc{Nemotron-9B} show partial movement, and the smallest two (\textsc{Olmo-3-7B}, \textsc{Qwen3-8B}) barely move. Retain-side movement follows the same ordering. Selectivity is model-determined rather than retriever-determined. \textsc{Llama-3.1-8B} loses between $51.2\%$ and $55.7\%$ of its retain-pool extractability across the three retrieval methods, while its forget side gains at most $28.9\%$. \textsc{Olmo-3-32B} is the only row where forget movement consistently exceeds retain movement. Switching from $\CorpusMid$ to $\CorpusPre$, or from BM25 to Infini-gram, reorders rows but does not change which models engage.

\begin{table*}[t]
\centering
\footnotesize
\setlength{\tabcolsep}{4pt}
\renewcommand{\arraystretch}{1.1}
\caption{Algorithm comparison at $|\ForgetReq|=50$ with EA curation (SimNPO, UNDIAL, RMU). Metrics and columns follow Table~\ref{tab:retrieval}.}
\label{tab:algos}
\begin{tabular}{ll|cc|c|cccccc|c}
\toprule
& & \multicolumn{2}{c|}{Net reduction (\%)} & QA  & \multicolumn{7}{c}{Validation $\Delta$ (pp)} \\
\cmidrule(lr){3-4} \cmidrule(lr){6-12}
& Model & F\,$\uparrow$ & R\,$\downarrow$ & $\Delta$ (pp) & GSM8K & BBH & WinoG & CoQA & HEval+ & LMBDA & \textit{Avg} \\
\midrule
\multirow{4}{*}{\rotatebox[origin=c]{90}{SimNPO}}
& \textsc{Llama-3.1-8B}   & \cellcolor{cg100} 100.0\% & \cellcolor{cr100} 100.0\% & \cellcolor{cr100} -5.7 & \cellcolor{cr100} -11.8 & \cellcolor{cr100} -37.6 & \cellcolor{cr75} -3.1 & \cellcolor{cr75} -3.9 & \cellcolor{cg75} +3.7 & \cellcolor{cr100} -10.1 & \cellcolor{cr100} -10.4 \\
& \textsc{Olmo-3-7B}      & \cellcolor{cg100} 100.0\% & \cellcolor{cr75} 72.2\% & \cellcolor{cr25} -1.0 & \cellcolor{cg50} +2.8 & \cellcolor{cr25} -0.1 & \cellcolor{cr50} -2.0 & \cellcolor{cr100} -5.4 & \cellcolor{cg50} +1.8 & \cellcolor{cr50} -2.3 & \cellcolor{cr25} -0.9 \\
& \textsc{Qwen3-8B}       & \cellcolor{cg100} 100.0\% & \cellcolor{cr100} 82.7\% & \cellcolor{cr100} -21.9 & \cellcolor{cr50} -3.0 & \cellcolor{cr50} -2.0 & \cellcolor{cg25} +0.3 & \cellcolor{cr100} -7.9 & \cellcolor{cr100} -4.9 & \cellcolor{cr100} -6.0 & \cellcolor{cr75} -3.9 \\
& \textit{Average}     & \cellcolor{cavg} 100.0\% & \cellcolor{cavg} 84.9\% & \cellcolor{cavg} -9.6 & \cellcolor{cavg} -4.0 & \cellcolor{cavg} -13.2 & \cellcolor{cavg} -1.6 & \cellcolor{cavg} -5.7 & \cellcolor{cavg} +0.2 & \cellcolor{cavg} -6.1 & \cellcolor{cavg} -5.1 \\
\midrule
\multirow{4}{*}{\rotatebox[origin=c]{90}{UNDIAL}}
& \textsc{Llama-3.1-8B}   & \cellcolor{cg100} 100.0\% & \cellcolor{cr100} 89.6\% & \cellcolor{cr50} -2.8 & \cellcolor{cg25} +1.4 & \cellcolor{cr25} -0.4 & \cellcolor{cr25} -0.3 & \cellcolor{cr100} -6.4 & \cellcolor{cg75} +3.7 & \cellcolor{cr50} -2.7 & \cellcolor{cr25} -0.8 \\
& \textsc{Olmo-3-7B}      & \cellcolor{cg100} 100.0\% & \cellcolor{cr100} 79.6\% & \cellcolor{cg25} +0.2 & \cellcolor{cr25} -0.8 & \cellcolor{cr25} -0.9 & \cellcolor{cr25} -0.9 & \cellcolor{cg100} +11.4 & \cellcolor{cg100} +5.5 & \cellcolor{cr25} -0.3 & \cellcolor{cg50} +2.3 \\
& \textsc{Qwen3-8B}       & \cellcolor{cg100} 100.0\% & \cellcolor{cr100} 83.1\% & \cellcolor{cr100} -5.2 & \cellcolor{cr50} -1.8 & \cellcolor{cg25} +0.7 & \cellcolor{cg25} +1.4 & \cellcolor{cr100} -5.4 & \cellcolor{cr75} -3.0 & \cellcolor{cg25} +0.9 & \cellcolor{cr25} -1.2 \\
& \textit{Average}     & \cellcolor{cavg} 100.0\% & \cellcolor{cavg} 84.1\% & \cellcolor{cavg} -2.6 & \cellcolor{cavg} -0.4 & \cellcolor{cavg} -0.2 & \cellcolor{cavg} +0.1 & \cellcolor{cavg} -0.1 & \cellcolor{cavg} +2.0 & \cellcolor{cavg} -0.7 & \cellcolor{cavg} +0.1 \\
\midrule
\multirow{4}{*}{\rotatebox[origin=c]{90}{RMU}}
& \textsc{Llama-3.1-8B}   & \cellcolor{cg100} 98.8\% & \cellcolor{cr100} 91.3\% & \cellcolor{cr50} -2.0 & \cellcolor{cr100} -10.7 & \cellcolor{cr100} -30.6 & \cellcolor{cr50} -1.6 & \cellcolor{cr100} -23.7 & \cellcolor{cg25} +1.2 & \cellcolor{cr100} -28.5 & \cellcolor{cr100} -15.6 \\
& \textsc{Olmo-3-7B}      & \cellcolor{cg100} 79.0\% & \cellcolor{cr100} 81.4\% & \cellcolor{cg25} +0.6 & \cellcolor{cg75} +3.8 & \cellcolor{cr100} -9.0 & \cellcolor{cg25} +0.4 & \cellcolor{cg100} +8.2 & \cellcolor{cg50} +2.4 & \cellcolor{cr100} -14.7 & \cellcolor{cr25} -1.5 \\
& \textsc{Qwen3-8B}       & \cellcolor{cg25} 24.5\% & \cellcolor{cr50} 25.7\% & \cellcolor{cr50} -2.3 & \cellcolor{cr50} -1.7 & \cellcolor{cr100} -5.6 & \cellcolor{cr50} -2.5 & \cellcolor{cg50} +1.6 & \cellcolor{cg100} +7.9 & \cellcolor{cr75} -3.8 & \cellcolor{cr25} -0.7 \\
& \textit{Average}     & \cellcolor{cavg} 67.5\% & \cellcolor{cavg} 66.1\% & \cellcolor{cavg} -1.2 & \cellcolor{cavg} -2.9 & \cellcolor{cavg} -15.0 & \cellcolor{cavg} -1.2 & \cellcolor{cavg} -4.6 & \cellcolor{cavg} +3.9 & \cellcolor{cavg} -15.7 & \cellcolor{cavg} -5.9 \\
\bottomrule
\end{tabular}
\end{table*}

\begin{table}[t]
\centering
\footnotesize
\setlength{\tabcolsep}{4pt}
\renewcommand{\arraystretch}{1.1}
\caption{Curator $\times$ unlearner cross evaluation at $|\ForgetReq|=50$. Each forget set is reused across unlearners, so only the unlearner varies within a curator block. Entries average \textsc{Llama-3.1-8B}, \textsc{Olmo-3-7B}, and \textsc{Qwen3-8B}; per-model results are in Appendix~\ref{app:cross} and Table~\ref{tab:algos}. Metrics follow Table~\ref{tab:retrieval}; these three-model means differ from its six-model averages.}
\label{tab:cross}
\begin{tabular}{ll|cc|c|cccccc|c}
\toprule
& & \multicolumn{2}{c|}{Net reduction (\%)} & QA & \multicolumn{7}{c}{Validation $\Delta$ (pp)} \\
\cmidrule(lr){3-4} \cmidrule(lr){6-12}
& Unlearner & F\,$\uparrow$ & R\,$\downarrow$ & $\Delta$ (pp) & GSM8K & BBH & WinoG & CoQA & HEval+ & LMBDA & \textit{Avg} \\
\midrule
\multirow{3}{*}{BM25 pre}
& SimNPO & \cellcolor{cg25} 10.5\% & \cellcolor{cr25} 18.7\% & \cellcolor{cr50} -1.5 & \cellcolor{cg25} +0.1 & \cellcolor{cg25} +0.5 & \cellcolor{cg25} +0.8 & \cellcolor{cr50} -1.8 & \cellcolor{cr50} -2.0 & \cellcolor{cg25} +0.9 & \cellcolor{cr25} -0.2 \\
& UNDIAL & \cellcolor{cg100} 98.2\% & \cellcolor{cr100} 83.5\% & \cellcolor{cr50} -1.6 & \cellcolor{cg25} +0.3 & \cellcolor{cr50} -1.8 & \cellcolor{cg25} +1.1 & \cellcolor{cr100} -7.7 & \cellcolor{cg25} +1.4 & \cellcolor{cr25} -0.9 & \cellcolor{cr25} -1.3 \\
& RMU & \cellcolor{cg50} 32.8\% & \cellcolor{cr50} 35.5\% & \cellcolor{cr50} -2.3 & \cellcolor{cr100} -16.1 & \cellcolor{cr100} -21.7 & \cellcolor{cr50} -2.2 & \cellcolor{cr100} -16.4 & \cellcolor{cg50} +1.8 & \cellcolor{cr100} -21.4 & \cellcolor{cr100} -12.7 \\
\midrule
\multirow{3}{*}{BM25 mid}
& SimNPO & \cellcolor{cg25} 6.8\% & \cellcolor{cr25} 20.1\% & \cellcolor{cr50} -2.3 & \cellcolor{cr25} -0.1 & \cellcolor{cg25} +0.3 & \cellcolor{cg25} +0.7 & \cellcolor{cr50} -1.6 & \cellcolor{cr75} -3.3 & \cellcolor{cg25} +0.6 & \cellcolor{cr25} -0.6 \\
& UNDIAL & \cellcolor{cg100} 92.9\% & \cellcolor{cr100} 86.5\% & \cellcolor{cr50} -2.5 & \cellcolor{cr50} -3.0 & \cellcolor{cr100} -10.9 & \cellcolor{cg25} +1.0 & \cellcolor{cr100} -15.0 & \cellcolor{cr25} -1.4 & \cellcolor{cr50} -2.3 & \cellcolor{cr100} -5.3 \\
& RMU & \cellcolor{cg50} 37.4\% & \cellcolor{cr50} 48.2\% & \cellcolor{cr100} -11.3 & \cellcolor{cr100} -18.1 & \cellcolor{cr100} -32.1 & \cellcolor{cr100} -6.9 & \cellcolor{cr100} -23.3 & \cellcolor{cr100} -6.9 & \cellcolor{cr100} -35.7 & \cellcolor{cr100} -20.5 \\
\midrule
\multirow{3}{*}{Infinigram mid}
& SimNPO & \cellcolor{cg25} 7.9\% & \cellcolor{cr25} 20.6\% & \cellcolor{cr50} -1.8 & \cellcolor{cg25} +0.1 & \cellcolor{cg25} +0.3 & \cellcolor{cg25} +1.0 & \cellcolor{cr25} -1.0 & \cellcolor{cr75} -3.5 & \cellcolor{cg25} +0.7 & \cellcolor{cr25} -0.4 \\
& UNDIAL & \cellcolor{cg100} 84.5\% & \cellcolor{cr100} 79.0\% & \cellcolor{cr50} -2.1 & \cellcolor{cg25} +0.2 & \cellcolor{cr75} -4.1 & \cellcolor{cg25} +1.1 & \cellcolor{cr100} -8.5 & \cellcolor{cg25} +1.2 & \cellcolor{cr50} -2.1 & \cellcolor{cr50} -2.0 \\
& RMU & \cellcolor{cg50} 34.8\% & \cellcolor{cr50} 44.3\% & \cellcolor{cr100} -7.3 & \cellcolor{cr100} -17.6 & \cellcolor{cr100} -28.5 & \cellcolor{cr100} -4.5 & \cellcolor{cr100} -19.8 & \cellcolor{cr25} -1.4 & \cellcolor{cr100} -33.8 & \cellcolor{cr100} -17.6 \\
\midrule
\midrule
\multirow{3}{*}{EA}
& SimNPO & \cellcolor{cg100} 100.0\% & \cellcolor{cr100} 84.9\% & \cellcolor{cr100} -9.6 & \cellcolor{cr75} -4.0 & \cellcolor{cr100} -13.2 & \cellcolor{cr50} -1.6 & \cellcolor{cr100} -5.7 & \cellcolor{cg25} +0.2 & \cellcolor{cr100} -6.1 & \cellcolor{cr100} -5.1 \\
& UNDIAL & \cellcolor{cg100} 100.0\% & \cellcolor{cr100} 84.1\% & \cellcolor{cr50} -2.6 & \cellcolor{cr25} -0.4 & \cellcolor{cr25} -0.2 & \cellcolor{cg25} +0.1 & \cellcolor{cr25} -0.1 & \cellcolor{cg50} +2.0 & \cellcolor{cr25} -0.7 & \cellcolor{cg25} +0.1 \\
& RMU & \cellcolor{cg75} 67.5\% & \cellcolor{cr75} 66.1\% & \cellcolor{cr25} -1.2 & \cellcolor{cr50} -2.9 & \cellcolor{cr100} -15.0 & \cellcolor{cr25} -1.2 & \cellcolor{cr100} -4.6 & \cellcolor{cg75} +3.9 & \cellcolor{cr100} -15.7 & \cellcolor{cr100} -5.9 \\
\bottomrule
\end{tabular}
\end{table}

\subsection{What if the target windows are given directly?}
\label{sec:exp-oracle}

EA drives forget extractability to near-100\% on every model, including the small models (\textsc{Olmo-3-7B}, \textsc{Qwen3-8B}) that retrieval barely moved (Table~\ref{tab:retrieval}). This shows that the models can be moved by direct target-window interventions under SimNPO; the crossed study below shows that retrieval outcomes cannot be attributed to the curator, model, or unlearner in isolation. Retain-pool extractability also falls on every row. We hypothesize that this reflects corpus support shared across works (Section~\ref{sec:cultural_embedding}). Capability cost varies by an order of magnitude across the suite, with damage concentrated on BBH and LAMBADA.

\subsection{Are evaluation aware curator's effects algorithm-specific?}
\label{sec:robustness-u}

To determine if the EA curator's collateral damage is specific to SimNPO, we evaluate two additional unlearning algorithms (UNDIAL and RMU) on \textsc{Llama-3.1-8B}, \textsc{Olmo-3-7B}, and \textsc{Qwen3-8B} (Table~\ref{tab:algos}). Across SimNPO and UNDIAL, $F$ reaches $100\%$ on all three models, and $R$ stays between 72--100\%. While both algorithms suffer from severe collateral forgetting by degrading retain-pool extractability, they produce very different capability outcomes on identical EA inputs. UNDIAL largely preserves average validation accuracy ($+0.1$ pp) while SimNPO degrades it (averaging $-5.1$ pp). RMU reduces $F$ on \textsc{Llama-3.1-8B} and \textsc{Olmo-3-7B}, with retain extractability falling alongside. On \textsc{Qwen3-8B}, both forget and retain stall together at $F{=}24.5\%$ and $R{=}25.7\%$, consistent with the hyperparameter sensitivity reported by~\citet{li2024wmdp}. The same retain-pool damage and capability gap persist at $|\ForgetReq|=100$ on \textsc{Llama-3.1-8B} and \textsc{Qwen3-8B}. The full table is in Appendix~\ref{app:scaling}. Table~\ref{tab:cross} repeats this comparison for the three retrieval curators, reusing an identical $\ForgetData$ within each $\langle$model, curator$\rangle$ pair so that only the unlearner changes. Holding the curator fixed, the unlearner changes the outcome dramatically: the same BM25-pre forget sets yield $10.5\%$ average forgetting under SimNPO, $98.2\%$ under UNDIAL, and $32.8\%$ under RMU, so weak suppression under SimNPO is not an intrinsic property of retrieval-derived forget sets. No configuration is selective, however: UNDIAL's strong suppression comes with $79.0$--$86.5\%$ retain-side suppression, per-model results in Appendix~\ref{app:cross}.


\paragraph{Takeaway.}
Retrieval-derived forget sets compose unpredictably with the downstream unlearner. Fixed-$\ForgetData$ algorithm comparisons remain valid, but their conclusions are conditional on the upstream curator; request-level comparisons evaluate the composed curator--unlearner pipeline. The evaluation-aware EA achieves forgetting on every model under SimNPO and UNDIAL, but the retain side falls and capability drops by up to $10.4$ pp on average. At the tested operating points, neither approach is selective. Lexical retrieval underspecifies the corpus support of a work because target text is widely shared across documents that are not the canonical source. Target-window indexing overspecifies it because the extractable windows for one work might be correlated with those for others. The same EA inputs produce a $5$-pp capability gap between SimNPO and UNDIAL, reinforcing that selective unlearning requires co-designing the curation strategy alongside the unlearning objective.

\section{Discussion and Future Direction}
\label{sec:discussion}
\paragraph{Unlearning needs curation.}
Our results suggest that the forget set should not be treated as a fixed premise of language-model unlearning. In realistic deployments, the request is often stated at the level of a work or behavior, while the unlearning algorithm requires concrete data to update against. What is selected for forgetting determines both whether the requested continuation becomes difficult to elicit and what else is disturbed. Forget set curation is therefore part of the unlearning problem, not merely a preprocessing detail.

\paragraph{Two insufficient endpoints.}
The experiments expose two natural but incomplete approaches to curation. Off-the-shelf corpus retrieval is not selective: lexical and exact-substring search can recover pieces of a work's corpus footprint, but the resulting forget sets compose unpredictably with the downstream unlearner and do not induce selective verbatim suppression at the tested operating points. Conversely, evaluation aware curation removes the retrieval bottleneck by selecting target windows directly, but still causes substantial output suppression of non-requested continuations and model-dependent capability regressions. Thus, the problem is not simply to find more target-like text. It is to construct a forget set whose effects remain localized after the model update.

\paragraph{Toward algorithm-aware curation.}
This localization depends on the downstream unlearning algorithm. The same evaluation aware forget set produces different capability profiles under SimNPO, UNDIAL, and RMU, suggesting that curation and unlearning should be evaluated jointly rather than as independent modules. Future curators may need to combine corpus-footprint signals with model-specific extraction profiles, dense or hybrid retrieval, influence estimation, or datamodeling to predict which examples will suppress the requested behavior without unnecessary collateral effects. \textsc{CleanSlate} studies a deliberately narrow setting: verbatim output suppression for songs and books. It does not address all forms of concept, entity, or factual unlearning, and our exact-overlap diagnostics miss paraphrase, translation, and semantic reuse. These limitations are also what make the task measurable. Extending forget set curation to richer request types, broader corpora, and algorithm-aware selection is a natural next step. The broader message is that practical unlearning should be studied end-to-end, from request, to curated forget set, to edited model.

\section*{Acknowledgments}
We thank Ludwig Schmidt, Etash Guha, Pratyush Maini, and Ananjan Nandi for helpful discussions. We acknowledge compute support by the Center for AI Safety. SK acknowledges support by NSF 2046795 and 2205329, IES R305C240046, ARPA-H, the MacArthur Foundation, Schmidt Sciences, HAI, OpenAI, Microsoft, and Google.

\bibliographystyle{plainnat}
\bibliography{references}

\newpage
\appendix
\part*{Appendix}

\section{Additional Related Work}
\label{app:additional_related_work}

\paragraph{Unlearning algorithms and benchmarks.}
A range of algorithms have been proposed for language-model unlearning. Gradient ascent on the forget loss is the simplest baseline. Negative preference optimization and SimNPO~\citep{zhang2024negative,fan2025simplicityprevailsrethinkingnegative} cast unlearning as preference optimization against a reference model. Representation Misdirection Unlearning~\citep{li2024wmdp} perturbs internal representations on forget data while regularizing retain representations, and UNDIAL~\citep{dong2025undial} adjusts logits through self-distillation. Several benchmarks evaluate whether such methods can remove specified behaviors while preserving utility. TOFU~\citep{maini2024tofu} provides synthetic author biographies, MUSE~\citep{shi2024muse} evaluates multiple aspects of language-model unlearning including verbatim memorization, WMDP~\citep{li2024wmdp} evaluates hazardous-knowledge unlearning, and OpenUnlearning~\citep{dorna2025openunlearning} provides a unified benchmarking framework. These benchmarks differ in domain and objective, but they largely evaluate the downstream unlearning step after the forget data, target examples, or target behaviors have already been specified.

\paragraph{Request-level, entity-level, and fact-level unlearning.}
Some recent work moves closer to settings where the target is specified at a higher level than a fixed forget corpus. RWKU~\citep{cao2024rwku} studies real-world knowledge unlearning, where the algorithm receives a target entity and the original model rather than an explicit training corpus. It then uses synthetic data produced by the model under evaluation to construct a forget set. This is close in spirit to our setting because the forget set is not directly provided, but the evaluated artifact differs: RWKU primarily measures unlearning algorithms under synthetic forget sets, while we evaluate the curation of forget sets under fixed unlearning algorithms. Deep fact unlearning~\citep{wu2024evaluating} studies whether a target fact remains inferable from retained facts under logical rules. This is related to our setting in that a target behavior can be supported by non-target evidence, but the unit and objective are different: deep unlearning targets factual deductive closure, while we study verbatim continuation and ask which corpus spans or documents should be used for suppression. 

\paragraph{Data for unlearning.}
Recent work shows that the contents of a forget set matter even after the forget corpus has been specified. \cite{pal2025llm} find that small subsets of benchmark-provided forget sets can match full-set unlearning. \cite{wan2025not} and \cite{zhou2025tokensmeantforgotten} show that token-level selection within known forget examples can reduce utility loss. \cite{allouah2026distributional} formalize distributional unlearning as selective data removal: given identified unwanted and retained sample sets, choose a small subset whose removal moves the edited data distribution away from the unwanted domain while preserving the retained one. These works are closely related to ours in emphasizing that selection matters, but they assume that the unwanted samples, retained samples, or target domain have already been obtained through an upstream process such as filtering, classification, or annotation. We study that upstream process itself. \citet{zhullm} identify the construction of forget sets as a major bottleneck in unlearning pipelines, and synthesize proxy forget data given only a broad domain name as input. This is close to our setting because it addresses the upstream dataset construction problem, but differs in focus, they work on synthesizing proxy data for broad domain-level forgetting, while we focus on measuring the role of the forget set (and the distributed corpus footprint) for suppressing verbatim output of specific targeted individual works. 

\section{Compute Requirements}
\label{appendix:compute}
All experiments were conducted on a compute node equipped with 8 NVIDIA H200 GPUs (141GB VRAM each), 230 CPU cores, 3TB of RAM, and 60TB of local NVMe storage. While model training utilized 1–2 GPUs, evaluation and validation tasks were performed on a single GPU. Notably, the corpus search and forget set curation phases are significantly memory- and storage-bound due to the scale of the datasets involved; these stages necessitated the full utilization of the available system memory and high-speed disk I/O.

\section{Search Corpora ($\Corpus$) Details}
\label{appendix:corpora_details}
For our experiments we use three distinct scale corpora
\begin{enumerate}[leftmargin=1.8em]
    \item \textbf{$\CorpusMid$ (Midtraining):} \texttt{allenai/dolma3\_dolmino\_mix-10B-1025}~\citep{olmo2025olmo} this is a midtraining mix which has coverage of  categories like instruction tuning data (FLAN, Tulu-3-SFT), code (\texttt{cranecode}, \texttt{stack\_edu}), and synthetic reasoning traces (Gemini, QwQ, Llama Nemotron, OpenThoughts). 
    \item \textbf{$\CorpusPre$ (Pretraining):} A $\sim11\%$ subset of Dolma-3 6T (\texttt{allenai/dolma3\_mix-6T}) pretraining mix~\citep{olmo2025olmo}. To make it computationally feasible, we limit this subset to five Common Crawl shards of the full pretraining corpus, \texttt{art\_and\_design}, \texttt{entertainment}, \texttt{history\_and\_geography}, \texttt{literature}, and \texttt{religion}.
    \item \textbf{$\CorpusCC$ (Web Scale):} January 2025 Common Crawl snapshot with $\sim9T$ tokens, accessed via the index provided by Infini-gram-mini~\citep{xu2025infini}. Used only for analysis of corpus footprint in~\cref{sec:cultural_embedding}
\end{enumerate}
\citet{olmo2025olmo} states December 2024 as the knowledge cutoff for $\CorpusMid$ and $\CorpusPre$.

\section{Computing Coverage of Songs and Books in Large Corpus}
\label{appendix:coverage_comp}
We compute these statistics using Infini-gram-mini~\citep{xu2025infini} but one can use any other dataset search tool. 
\subsection{Computing $N-$Gram Matches and Coverage}
\label{appendix:cultural_embedding:algo}
To quantify the verbatim overlap between a reference sequence and a large-scale corpus $\Corpus$, we employ an iterative retrieval algorithm that identifies the maximal exact $n$-gram matches starting at every word start position. Let $T_w=(x_1,\ldots,x_m)$ be the whitespace-delimited word sequence of work $w$. For start position $i$ and length $n$, let $s_{i,n}$ be the exact character span covering $x_i,\ldots,x_{i+n-1}$, preserving punctuation and spacing. For indexed corpus $\mathcal{C}$, we define the frequency $C_{\Corpus}(s)$ and the coverage statistic $\mathrm{cov}_{\geq N}(w;\Corpus)$ as follows:
\begin{equation}
\begin{aligned}
C_{\Corpus}(s) &= \sum_{d\in\Corpus}\#_d(s) \\
\mathrm{cov}_{\geq N}(w;\Corpus) &= \frac{1}{m}\sum_{j=1}^{m} \mathbf{1}\!\left[\,\exists\,i,n:\ n\geq N,\ i\leq j<i+n,\ C_{\Corpus}(s_{i,n})>0\,\right]
\end{aligned}
\end{equation}
The statistic $\mathrm{cov}_{\geq N}$ measures the fraction of word positions in $w$ covered by at least one verbatim $n$-gram of length $n\geq N$ present in $\Corpus$. For each start position $i$, we begin at $n_{\min}=5$ and extend $n$ while $C_{\Corpus}(s_{i,n})>0$. Short $n$-grams ($n < 5$) occur with high background frequency due to linguistic coincidence. 

\subsection{Aggregate Coverage Statistics}
\label{appendix:cultural_embedding:aggregate}
 Out of $4{,}663$  works in \benchmark, $4{,}596$ ($98.6\%$) retrieve at least one positive-count 5-gram match with documents in $\CorpusMid$. The distribution of retrieved documents per work exhibits a median of $119$, a $90$th percentile of $299$, and a maximum of $45{,}338$. For source inspection in $\CorpusMid$ we sample $D=20$ documents for each of the top-K $(K=20000)$ spans ranked by length and occurrence count, for $\CorpusCC$ we sample $D=2$ documents for each of the top $K=2000$ spans, due to computational limitations.
 \begin{figure}
    \centering
    \includegraphics[width=\linewidth]{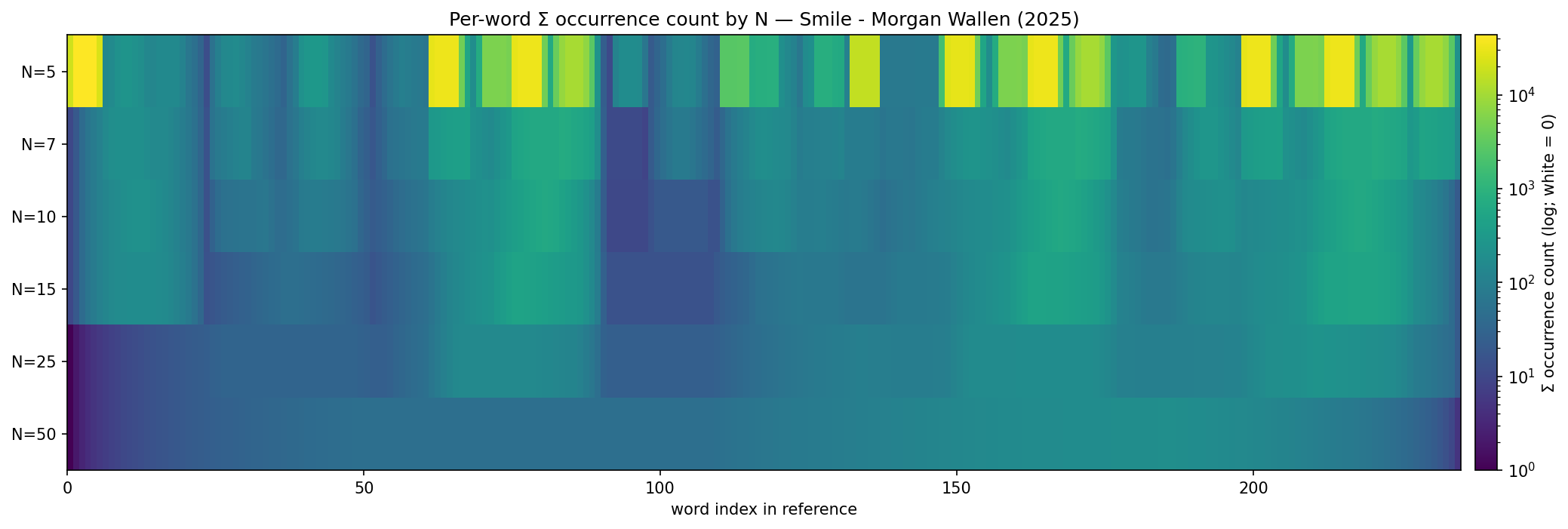}
    \caption{N-gram coverage for \emph{Smile} in $\CorpusCC$}
    \label{fig:coverage_smile}
\end{figure}

\subsection{Extractability vs. Footprint Density Plots}
\label{appendix:cultural_embedding:plots}
To understand the correlation between localized corpus prevalence and model extractability, we plot the \emph{footprint density} at a given word position $j$, alongside the maximum extraction probability ($p_z$) measured amongst all suffixes $z$ covering the position $j$. The density aggregates the occurrence counts of all valid $n$-grams in $\CorpusMid$ that overlap the $j$-th word of the work.

\begin{figure}
    \centering
    \includegraphics[width=\linewidth]{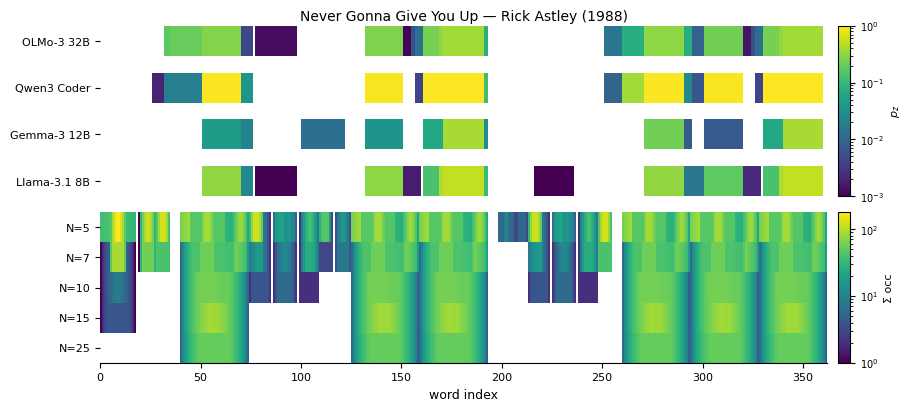}
    \caption{Footprint density (sum of matches of all overlapping $n$-gram matches in $\CorpusMid$) vs. maximum extraction probability ($p_z$) per word for \emph{Never Gonna Give You Up}.}
    \label{fig:mem_coverage_song_nggyu}
\end{figure}

\begin{figure}
    \centering
    \includegraphics[width=1\linewidth]{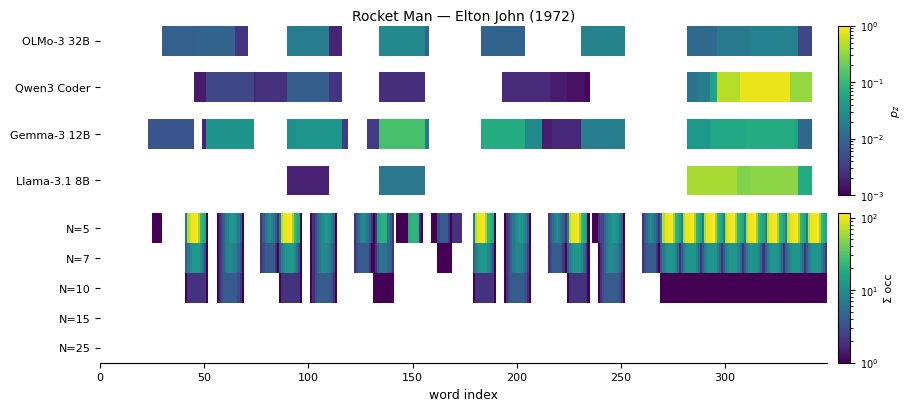}
     \caption{Footprint density vs. maximum extraction probability ($p_z$) per position for \emph{Rocket Man}.}
    \label{fig:mem_coverage_song_rocket_man}
\end{figure}

\begin{figure}
    \centering
    \includegraphics[width=1\linewidth]{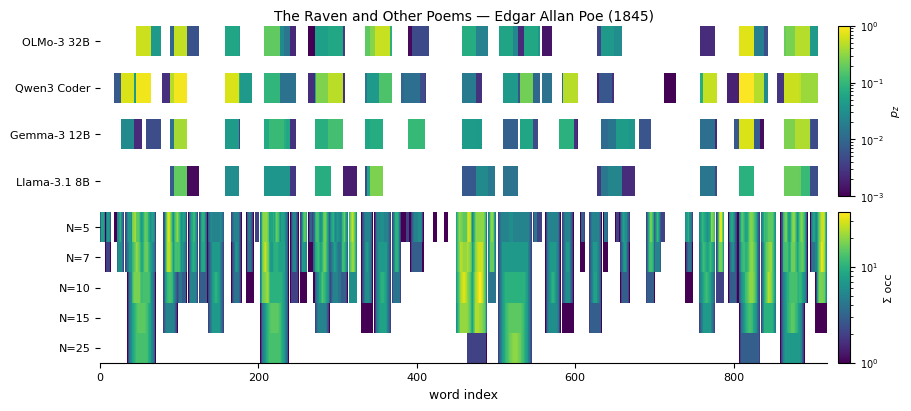}
    \caption{Footprint density in $\CorpusMid$ vs. maximum extraction probability ($p_z$) per position for poems of Edgar Allan Poe}
    \label{fig:mem_coverage_poem}
\end{figure}

\begin{figure}
    \centering
    \includegraphics[width=1\linewidth]{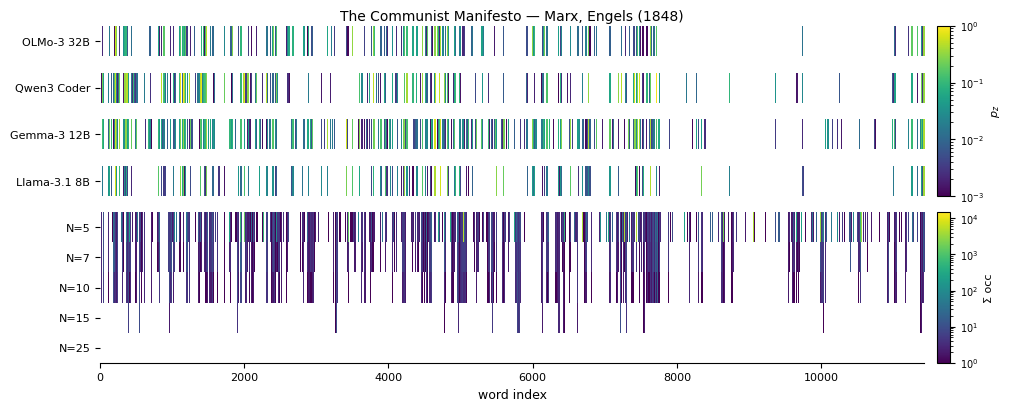}
     \caption{Footprint density in $\CorpusMid$ vs. max$p_z$ per position for \emph{The Communist Manifesto}.}
    \label{fig:mem_coverage_communist_manifesto}
\end{figure}

Figures~\ref{fig:mem_coverage_song_nggyu} through \ref{fig:mem_coverage_communist_manifesto} corroborate that peaks in $\CorpusMid$ occurrences strongly align with spikes in extractability. For instance, the choruses of popular songs (\emph{Never Gonna Give You Up}, \emph{Rocket Man}) and famous refrains in poems (\emph{A Dream Within a Dream}) exhibit massive frequency spikes in the corpus. The model's extraction probability neatly mirrors these spikes, rising precisely where the footprint density is highest.

\clearpage
\section{Baseline Extractability Patterns Across Model Families} 
\label{appendix:memorization_flame}

Extractability is neither uniform across target works nor consistent across model architectures. Table~\ref{tab:baseline} documents the baseline characteristics of our evaluation suite prior to any curation or unlearning. The extraction threshold ($p_z \geq 0.001$) is evaluated over sliding windows of $100$ prefix  and $100$ suffix characters. 

\begin{figure}
    \centering
    \includegraphics[width=1\linewidth]{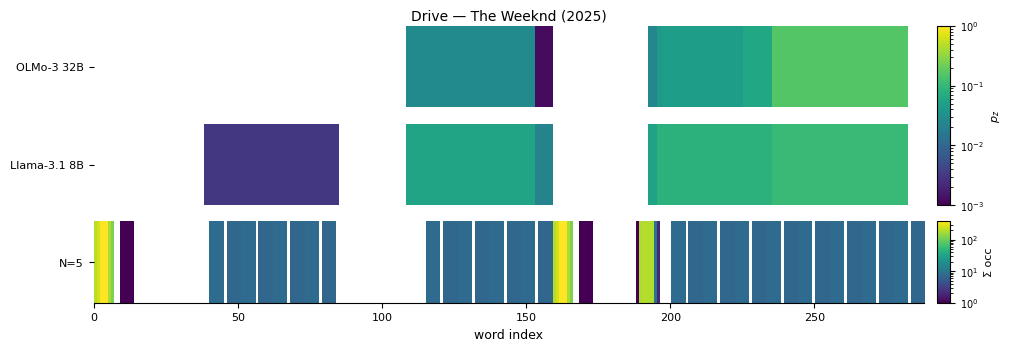}
    \caption{$\max p_z$ across \textit{Drive} by Weekend, both models pre date the song, windows with high $p_z$ are highly repetitive portions of the chorus}
    \label{fig:drive_p_z}
\end{figure}

\paragraph{Model scale and capability.} 
Larger models consistently exhibit higher raw extractability. For instance, the 32B-parameter \textsc{Olmo-3-32B} achieves a forget-pool extraction rate (Ext-F) of 8.27\%, while its 7B-parameter counterpart (\textsc{Olmo-3-7B}) achieves 7.20\%. Base models also tend to exhibit higher extractability than their instruction-tuned variants, probably due to alignment training penalizing raw regurgitation in favor of conversational formatting. 
Repetitive structure of certain songs also increase their extractability, \textit{Drive} by The Weeknd (2025) post dates Llama-3.1-8B and Olmo-32b but both model assign high $p_z$  where a sections of the work are near identical prefix–suffix pair (see Figure~\ref{fig:drive_p_z}), this suggests that the language-modeling objective by itself can increase the probability of extraction for highly repetitive works.
\paragraph{Heterogeneity of extractable text.} 
The data reveals that extractability is highly localized within the texts themselves. For songs, verses often fall below the extraction threshold, while choruses reinforced by both their internal repetition and their higher external footprint density cross the threshold easily. More importantly, \emph{different models extract different windows}. A specific verse that is highly extractable for \textsc{Llama-3.1-8B} may fall below the threshold for \textsc{Nemotron-9B} due to differences in their respective training mixtures. 

\paragraph{QA performance vs. verbatim extraction.} 
Table~\ref{tab:baseline} also reports the baseline \textsc{CleanSlate}-QA accuracy (ranging from 22.01\% to 35.07\% pass@5). The fact that these models can reliably answer factual questions about the content confirms that they possess abstract knowledge of the works. The core challenge of forget set curation is to selectively suppress the localized $p_z$ spikes responsible for exact extraction while leaving this broader semantic knowledge (QA) and general capabilities (GSM8K, etc.) undisturbed.

\begin{table*}[t]
\centering
\tiny
\setlength{\tabcolsep}{3pt}
\setlength{\tabcolsep}{5pt}
\renewcommand{\arraystretch}{1.1}
\caption{Baseline evaluation models prior to any curation or unlearning, ordered by forget-pool extractability. The \textit{Forget pool} and \textit{Retain pool} blocks report the total number of $(100, 100)$ character windows ($N$) and the number that meet the extraction threshold $p_z \geq 0.001$ (\textit{Ext.}). \textit{QA} is CleanSlate-QA pass@5 accuracy (\%) averaged over forget and retain items. Remaining columns report baseline accuracy (\%) on the validation suite. Forget pool size is $|\mathcal{F}|{=}50$ except for Olmo-3-7B, which has $|\mathcal{F}|{=}38$. GSM8K uses flexible-extract scoring.}
\label{tab:baseline}
\begin{tabular}{l|cc|cc|c|ccccccc}
\toprule
& \multicolumn{2}{c|}{Forget pool} & \multicolumn{2}{c|}{Retain pool} & QA (\%) & \multicolumn{7}{c}{Validation accuracy (\%)} \\
\cmidrule(lr){2-3} \cmidrule(lr){4-5} \cmidrule(lr){7-13}
Model & $N$ & Ext. & $N$ & Ext. & & GSM8K & BBH & HSwag & WinoG & CoQA & HEval+ & LMBDA \\
\midrule
Llama-3.1-8B  & 8{,}513  & 599   & 3{,}999{,}813 & 11{,}130 & 25.63 & 50.64 & 64.57 & 60.69 & 74.43 & 49.10 & 31.71 & 76.29 \\
Olmo-3-7B     & 8{,}084  & 582   & 4{,}000{,}242 & 2{,}108  & 27.31 & 69.60 & 61.63 & 54.13 & 69.77 & 57.63 & 31.71 & 69.63 \\
Nemotron-9B   & 8{,}840  & 647   & 3{,}999{,}486 & 3{,}082  & 22.01 & 76.80 & 72.67 & 58.11 & 73.48 & 53.92 & 10.98 & 68.00 \\
Qwen3-8B      & 10{,}128 & 755   & 3{,}998{,}198 & 3{,}297  & 34.71 & 88.55 & 79.48 & 57.12 & 67.88 & 64.78 & 58.54 & 65.17 \\
Gemma-3-12B   & 27{,}936 & 2{,}210 & 3{,}980{,}390 & 14{,}501 & 30.63 & 71.80 & 73.91 & 62.04 & 75.30 & 50.03 & 12.80 & 76.03 \\
Olmo-3-32B    & 67{,}323 & 5{,}565 & 3{,}941{,}003 & 17{,}280 & 35.07 & 78.24 & 78.73 & 60.97 & 76.48 & 58.38 & 39.63 & 76.31 \\

\bottomrule
\end{tabular}
\end{table*}



\section{CleanSlate-QA Benchmark Construction}
\label{app:qa_construction}

Here we go over the pipeline used to build CleanSlate-QA, the content-grounded retain-metric benchmark referenced in Section~\ref{sec:qa} and reported as the \textit{QA} column of Table~\ref{tab:baseline}. Construction has three stages. Stage 1 extracts factual propositions from each work, stage 2 turns each proposition into an atomic question-answer pair, and stage 3 filters candidates by out-of-sample (OOS) model knowledge.

\paragraph{Routing and infrastructure.} Each work in $\Works$ is routed by length. Books are chunked into $60{,}000$-character windows that are extracted while songs are extracted in a single call. Stages 1 and 2 dispatch \texttt{gemini-3.1-flash-lite-preview}. Stage 3 uses (\texttt{Qwen/Qwen3.5-9B}, \texttt{meta-llama/Meta-Llama-3-8B-Instruct-Lite}).

\subsection{Stage 1: Proposition Extraction}

The model is asked to extract a small set of \emph{interior factual propositions}, defined as statements anchored to a named entity, place, number, or concrete event mentioned inside the text, with priors such as title, creator, year, and genre explicitly excluded. Each proposition must cite a short source span from the work. The song and book prompts are reproduced below.

\begin{tcolorbox}[
    colback=gray!5!white,
    colframe=gray!30,
    title=\textbf{\textcolor{black}{Song Proposition Extraction Prompt}},
    fonttitle=\normalsize,
    breakable
]
\small
You are extracting interior facts from a song's lyrics that will be used to build a retain-metric QA benchmark.

\medskip
Title: ``\verb|{title}|''\\
Creator: \verb|{creator}|

\medskip
LYRICS:\\
\verb|{text}|

\medskip
\textbf{TASK:} Extract between 3 and 8 \textbf{interior factual propositions} from these lyrics.

\medskip
A good proposition:
\begin{itemize}
    \item Refers to a named entity, specific detail, number, place, person, action, or relationship mentioned INSIDE the lyrics.
    \item Is a full factual statement, not just a word.
    \item Cites a short source span (1--2 lines from the lyrics) where the fact appears.
\end{itemize}

A bad proposition:
\begin{itemize}
    \item Is about the title, creator/artist, year, genre (priors).
    \item Restates the hook, chorus, or title in different words (e.g.\ for ``Ladies' Night'': ``the song says it is your night'' --- that's just the hook).
    \item Has no named entity, place, number, or specific concrete detail.
    \item Is a generic feeling, mood, theme, or exhortation (``the song tells listeners to dance'').
    \item Uses pronouns/possessives as its key content (``your'', ``their'', ``this'').
\end{itemize}

If the song is abstract with few concrete details, return fewer propositions. Do NOT pad.

\medskip
Return ONLY a JSON object:\\
\texttt{\{"propositions": [\{"fact": "...", "source\_span": "..."\}, ...]\}}
\end{tcolorbox}

\begin{tcolorbox}[
    colback=gray!5!white,
    colframe=gray!30,
    title=\textbf{\textcolor{black}{Book Proposition Extraction Prompt (per chunk)}},
    fonttitle=\normalsize,
    breakable
]
\small
You are extracting interior facts from a CHUNK of a book that will be used to build a retain-metric QA benchmark.

\medskip
Book: ``\verb|{title}|''\\
Author: \verb|{creator}|\\
Chunk \verb|{chunk_idx}| of \verb|{total_chunks}|

\medskip
TEXT:\\
\verb|{text}|

\medskip
\textbf{TASK:} Extract between 5 and 15 \textbf{interior factual propositions} from this chunk.

\medskip
Prioritize:
\begin{itemize}
    \item Named characters and their distinguishing features, relationships, actions.
    \item Specific locations, settings, objects mentioned.
    \item Concrete plot events in this chunk.
    \item Numeric specifics (ages, dates within the narrative, counts).
\end{itemize}

Avoid:
\begin{itemize}
    \item Anything about the title, author, publication year, genre (priors).
    \item Generic theme/mood statements without a concrete anchor.
    \item Verbatim famous quotes unless they reveal a concrete fact.
\end{itemize}

Each proposition must cite a short source span (1--3 lines) from the chunk.

\medskip
Return ONLY a JSON object:\\
\texttt{\{"propositions": [\{"fact": "...", "source\_span": "..."\}, ...]\}}
\end{tcolorbox}

\subsection{Stage 2: Atomic QA Generation}

Each proposition is converted into one atomic question-answer pair. Questions must be self-contained (naming both title and creator), target an interior detail rather than a prior, and admit a short 1--5 word answer that is itself a named entity, number, specific place, specific object, or proper noun. The model is allowed to return an empty list when no valid pair can be produced for a given proposition. This is the primary mechanism by which low-yield propositions are dropped.

\begin{tcolorbox}[
    colback=gray!5!white,
    colframe=gray!30,
    title=\textbf{\textcolor{black}{Atomic QA Generation Prompt}},
    fonttitle=\normalsize,
    breakable
]
\small
Turn this fact into ONE atomic question--answer pair for a retain-metric benchmark.

\medskip
Title: ``\verb|{title}|''\\
Creator: \verb|{creator}|\\
Fact: \verb|{fact}|\\
Source span: \verb|{source_span}|

\medskip
\textbf{REQUIREMENTS} --- the question must:
\begin{enumerate}
    \item Be SELF-CONTAINED (atomic): name the title AND creator naturally. The model sees only the question, no context.
    \item Have a SHORT semantic answer (1--5 words) that is a \textbf{named entity, number, specific place, specific object, or proper noun}. Not a pronoun, possessive, or generic colloquial phrase.
    \item Target an interior detail --- NOT the title, creator, year, or genre.
    \item The answer must NOT be the title or any substring of the title/creator.
    \item The answer must NOT appear as a phrase inside your own question text.
    \item Not be answerable from the title alone (e.g., if the song is ``Sexy + 17'' don't ask the girl's age).
\end{enumerate}

\medskip
GOOD examples:
\begin{itemize}
    \item Q: ``In Dr.\ Hook's `Sylvia's Mother', how does the caller address Sylvia's mother?'' \quad A: ``Mrs.\ Avery''
    \item Q: ``In `C'mon N' Ride It' by Quad City DJ's, what car does the narrator want to be in the back of?'' \quad A: ``Impala''
\end{itemize}

BAD examples:
\begin{itemize}
    \item Q: ``Who sang `Hotline Bling'?'' $\rightarrow$ priors
    \item Q: ``Complete: `Ride that choo-choo \_\_\_'\,'' $\rightarrow$ verbatim
    \item Q: ``What is the mood of `More Than A Woman'?'' $\rightarrow$ subjective
    \item Q: ``In Kool \& the Gang's `Ladies' Night', what does the song say is happening tonight?'' A: ``your night'' $\rightarrow$ hook-echo, pronoun answer, no specific content
\end{itemize}

\medskip
If the fact cannot produce a good atomic Q meeting all requirements, return an empty list.

\medskip
Return ONLY a JSON object shaped like:\\
\texttt{\{"pairs": [\{"question": "...", "answer": "..."\}, ...]\}}
\end{tcolorbox}

\subsection{Stage 3: OOS Knowledge Filter and Judge}

To drop questions whose answers can only be recovered by memorizing the source work, each candidate question is presented to the OOS probe models with no surrounding context. A candidate is kept iff at least one probe model produces a correct response.

\begin{tcolorbox}[
    colback=gray!5!white,
    colframe=gray!30,
    title=\textbf{\textcolor{black}{LLM Judge Prompt}},
    fonttitle=\normalsize,
    breakable
]
\small
Does the model answer contain the reference answer or convey the same meaning?

\medskip
Reference: \verb|{reference}|\\
Model Answer: \verb|{model_answer}|

\medskip
Reply with ONLY this JSON, nothing else: \texttt{\{"correct": true\}} or \texttt{\{"correct": false\}}
\end{tcolorbox}

\section{Unlearning Training Details}
\label{sec:unlearning_hparams}
The trainer consumes curated forget rows of the form $(x,z)$, where $x$ is a prefix and $z$ is the suffix to suppress.  Each training item is anchored on one forget example and paired with a randomly sampled retain example. The total loss is
\[
    \lambda_f \mathcal{L}_{\mathrm{forget}}
    +
    \lambda_r \mathcal{L}_{\mathrm{retain}} .
\]
For SimNPO we use the average suffix NLL form
\[
    \mathcal{L}_{\mathrm{SimNPO}}
    =
    -\frac{2}{\beta}
    \log \sigma\!\left(
        \beta(\bar{\ell}_\theta(z\mid x)-\delta)
    \right),
\]
with $\mathcal{L}_{\mathrm{retain}}$ equal to retain NLL. UNDIAL distills from a frozen reference model after subtracting $\beta_{\mathrm{U}}$ from the gold-token teacher logit on forget suffix tokens. RMU minimizes MSE between forget activations and a random control vector at decoder block \texttt{model.layers.7}, and uses an activation matching retain loss against the frozen reference model. Table~\ref{tab:hyperparams} captures all hyperparameters specific to $\Unlearner$.
\begin{table}
\centering
\small
\caption{Unlearning hyperparameters.}
\label{tab:hyperparams}
\begin{tabular}{lccc}
\toprule
& SimNPO & UNDIAL & RMU \\
\midrule
$\beta$ / $\delta$ & 2.0 / 0.0 & -- & -- \\
$\beta_{\mathrm{U}}$ & -- & 10.0 & -- \\
RMU coeff. / layer & -- & -- & 2.0 / \texttt{layers.7} \\
$\lambda_f$ / $\lambda_r$ & 0.125 / 1.0 & 1.0 / 1.0 & 1.0 / 1.0 \\
Retain loss & NLL & NLL & embed-diff \\
Epochs & 1 & 2 & 2 \\
LR / warmup & $10^{-5}$ / 10 & $10^{-5}$ / 20 & $10^{-5}$ / 20 \\
Batch / accum. & 8 / 4 & 8 / 4 & 8 / 4 \\
\bottomrule
\end{tabular}
\end{table}

\subsection{Some example QA pairs}
\begin{figure}
    \centering
    \includegraphics[width=1\linewidth]{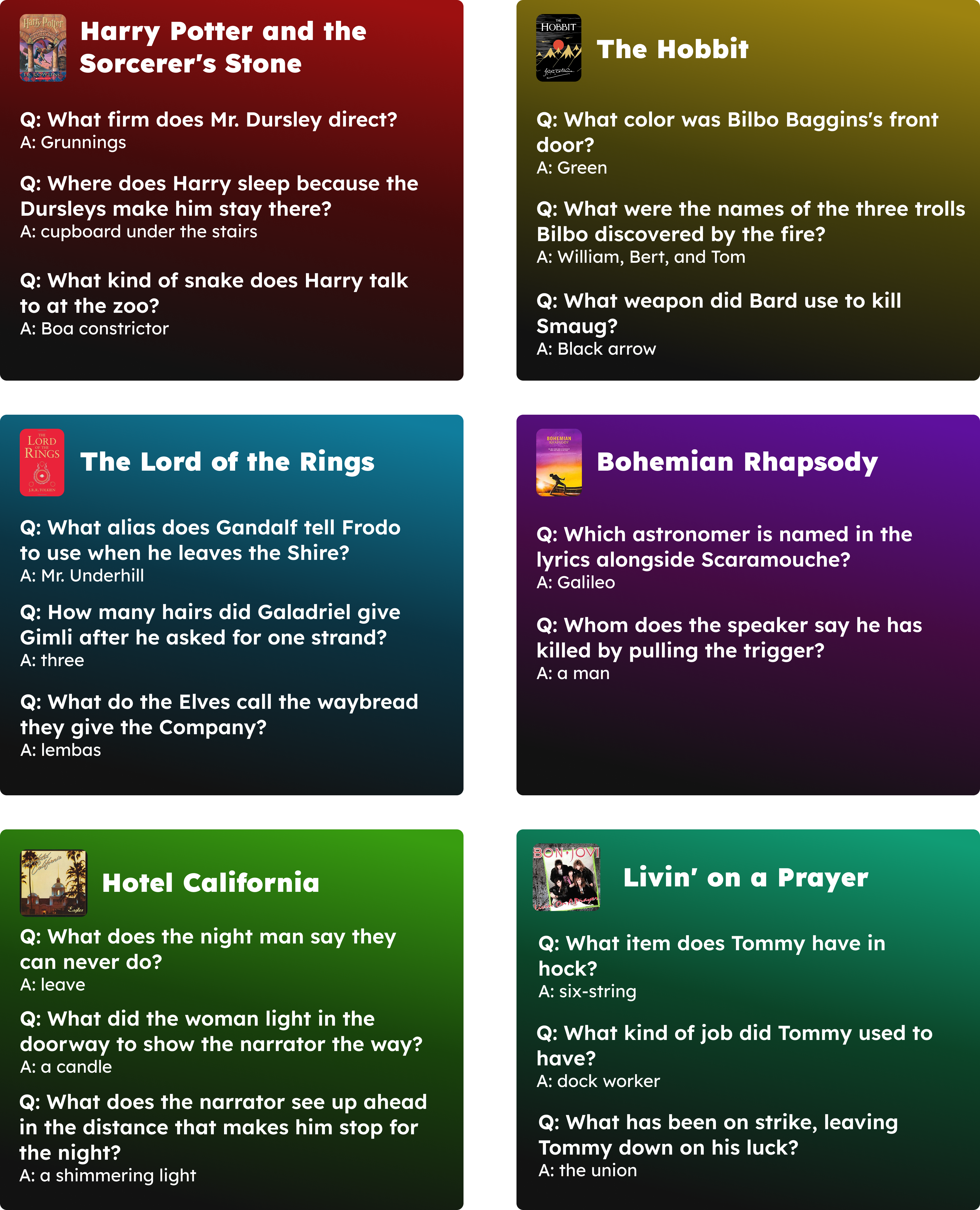}
     \caption{Example QA pairs from \benchmark}
    \label{fig:QACLEANSLATE}
\end{figure}

See Figure~\ref{fig:QACLEANSLATE}

\section{Retrieval Curators and Forget-Set Construction Details}
\label{app:retrieval_details}
Each retrieval curator maps a request text to a set of prefix--suffix training examples in three steps: retrieve corpus units, project retrieved text into
evaluation-shaped windows, and allocate a bounded number of windows per requested work. The unlearning trainer consumes only these pre-sliced rows: it tokenizes the concatenated prefix and suffix and masks prefix tokens from the forget loss.

\paragraph{BM25 retrieval.}
For BM25, we build sharded BM25s indices over the search corpus. Documents shorter than 100 characters are discarded. Remaining documents are split at word boundaries into segments of at most 2{,}000 characters with 500-character overlap; segments shorter than 200 characters are removed. BM25 tokenization lowercases text and removes English stopwords. At search time, each shard returns its top candidates and we retain the global top 100 segments for the requested work. Very long request texts are queried through sampled probes, after which the same global top-$k$ merge is applied.

\paragraph{Infini-gram retrieval.}
We use an Infini-gram-mini over the corpus. For each character offset $i$ in the request text $T$, we first test whether the 20-character substring beginning at $i$ occurs in the corpus. If it does, similar to OlmoTrace~\cite{liu2025olmotrace} we binary search for the longest substring $T[i:j]$ with positive corpus count and record the span together with its count. After scanning all offsets, we  rank the spans by character length, and keep the top 100. For each retained span, the retrieval pipeline queries corpus contexts containing the span; the curation artifact stores the first retrieved context, while metadata records the matched \texttt{span\_text}, corpus count, coverage, and number of retrieved contexts.

\paragraph{Window extraction.}
We use the same character window shape for all forget sets. Let $W=100$ and $S=10$. A window beginning at character position $s$ emits
\[
    \mathrm{prefix}=D[s:s+W],
    \qquad
    \mathrm{suffix}=D[s+W:s+2W].
\]
EA curator applies this sliding window directly to the reference text. BM25 applies it to the retrieved BM25 text unit recorded in the retrieval artifact. Infini-gram based curator first locates \texttt{span\_text} inside the retrieved corpus context and emits stride-spaced windows whose 200-character frame contains the midpoint of the matched span. Duplicate prefix- suffix pairs are removed within each requested work.

\paragraph{Quota allocation.}
Retrieval can produce many overlapping windows from a small number of documents. To avoid letting a single repeated source dominate the forget set, we cap each requested work at $C=128$ windows and allocate this budget across retrieved documents. Let $v_i$ be the number of available windows for document $i$, ordered by retrieval rank. We first give each document up to $F=4$ windows, bounded by availability and the remaining budget. Any remaining budget is then distributed round-robin over documents in rank order until the cap is reached or no document has unused windows. If a document has more available windows than its quota, we take evenly spaced windows from that document.

\paragraph{Retain data.}
Following prior work~\citep{gandikota2024erasing,li2024wmdp} we use \texttt{WikiText} as the retain set $\RetainData$, we keep this fixed across curators to isolate the effect of the forget set $\ForgetData$. During training, each forget example is paired with a randomly sampled retain example when the unlearning objective uses retain regularization.

\section{Scaling Forget Request}
\label{app:scaling}
See Table~\ref{tab:poolsize}.

\begin{table*}[!h]
\centering
\footnotesize
\setlength{\tabcolsep}{4pt}
\renewcommand{\arraystretch}{1.1}
\caption{Forget-set size ablation at $|\mathcal{F}|=100$, reported for Llama-3.1-8B and Qwen3-8B. Metrics and columns follow Table~\ref{tab:retrieval}.}
\label{tab:poolsize}
\begin{tabular}{ll|cc|c|cccccc|c}
\toprule
& & \multicolumn{2}{c|}{Net reduction (\%)} & QA & \multicolumn{7}{c}{Validation $\Delta$ (pp)} \\
\cmidrule(lr){3-4} \cmidrule(lr){6-12}
& Method & F\,$\uparrow$ & R\,$\downarrow$ & $\Delta$ (pp) & GSM8K & BBH & WinoG & CoQA & HEval+ & LMBDA & \textit{Avg} \\
\midrule
\multirow{5}{*}{\rotatebox[origin=c]{90}{Llama-3.1-8B}}
& EA (UNDIAL)      & \cellcolor{cg100} 100.0\% & \cellcolor{cr100} 93.0\% & \cellcolor{cr75} -4.0 & \cellcolor{cg25} +1.2 & \cellcolor{cr25} -0.6 & \cellcolor{cr50} -1.8 & \cellcolor{cr100} -4.5 & \cellcolor{cg50} +2.4 & \cellcolor{cr75} -4.0 & \cellcolor{cr25} -1.2 \\
& EA (SimNPO)      & \cellcolor{cg100} 100.0\% & \cellcolor{cr100} 99.4\% & \cellcolor{cr100} -5.9 & \cellcolor{cr50} -2.8 & \cellcolor{cr50} -2.4 & \cellcolor{cr75} -3.4 & \cellcolor{cg75} +3.7 & \cellcolor{cg75} +3.0 & \cellcolor{cr100} -7.0 & \cellcolor{cr25} -1.5 \\
& EA (RMU)         & \cellcolor{cg100} 99.5\% & \cellcolor{cr100} 94.0\% & \cellcolor{cr50} -1.7 & \cellcolor{cr100} -25.0 & \cellcolor{cr100} -58.0 & \cellcolor{cr50} -1.7 & \cellcolor{cr100} -48.7 & \cellcolor{cg50} +1.8 & \cellcolor{cr100} -45.6 & \cellcolor{cr100} -29.5 \\
& Infinigram mid       & \cellcolor{cg50} 49.0\% & \cellcolor{cr75} 70.8\% & \cellcolor{cr50} -2.5 & \cellcolor{cr25} -0.9 & \cellcolor{cr25} -1.4 & \cellcolor{cr50} -1.6 & \cellcolor{cg50} +2.1 & \cellcolor{cg50} +1.8 & \cellcolor{cr50} -2.6 & \cellcolor{cr25} -0.4 \\
& BM25 mid             & \cellcolor{cg50} 38.8\% & \cellcolor{cr75} 64.9\% & \cellcolor{cr50} -2.3 & \cellcolor{cr50} -1.9 & \cellcolor{cr25} -1.2 & \cellcolor{cr50} -1.7 & \cellcolor{cg100} +5.0 & \cellcolor{cg75} +4.3 & \cellcolor{cr50} -2.8 & \cellcolor{cg25} +0.3 \\
\midrule
\multirow{5}{*}{\rotatebox[origin=c]{90}{Qwen3-8B}}
& EA (UNDIAL)      & \cellcolor{cg100} 100.0\% & \cellcolor{cr100} 88.0\% & \cellcolor{cr100} -6.9 & \cellcolor{cr50} -2.7 & \cellcolor{cr25} -0.7 & \cellcolor{cr25} -0.1 & \cellcolor{cr75} -3.9 & \cellcolor{cg25} +0.6 & \cellcolor{cr25} -1.0 & \cellcolor{cr25} -1.3 \\
& EA (SimNPO)      & \cellcolor{cg100} 94.2\% & \cellcolor{cr75} 72.8\% & \cellcolor{cr100} -10.9 & \cellcolor{cr25} -1.2 & \cellcolor{cg50} +2.0 & \cellcolor{cg50} +2.1 & \cellcolor{cr100} -4.7 & \cellcolor{cr75} -3.0 & \cellcolor{cg25} +1.1 & \cellcolor{cr25} -0.6 \\
& EA (RMU)         & \cellcolor{cg50} 45.8\% & \cellcolor{cr75} 50.0\% & \cellcolor{cr100} -4.6 & \cellcolor{cr50} -2.6 & \cellcolor{cr100} -7.8 & \cellcolor{cr50} -1.7 & \cellcolor{cr25} -1.3 & \cellcolor{cg100} +4.9 & \cellcolor{cr100} -6.1 & \cellcolor{cr50} -2.4 \\
& Infinigram mid       & \cellcolor{cg25} 4.4\% & \cellcolor{cr25} 3.4\% & \cellcolor{cr100} -6.6 & \cellcolor{cr50} -1.7 & \cellcolor{cg50} +1.5 & \cellcolor{cg50} +2.9 & \cellcolor{cr50} -1.6 & \cellcolor{cr100} -9.1 & \cellcolor{cg75} +3.4 & \cellcolor{cr25} -0.8 \\
& BM25 mid             & \cellcolor{cg25} 4.8\% & \cellcolor{cr25} 3.9\% & \cellcolor{cr100} -6.7 & \cellcolor{cr25} -1.3 & \cellcolor{cg25} +1.4 & \cellcolor{cg50} +2.4 & \cellcolor{cr50} -2.2 & \cellcolor{cr100} -5.5 & \cellcolor{cg75} +3.3 & \cellcolor{cr25} -0.3 \\
\bottomrule
\end{tabular}
\end{table*}

\section{Curator \texorpdfstring{$\times$}{x} Unlearner \texorpdfstring{$\times$}{x} Model Cross Evaluation}
\label{app:cross}

{See Table~\ref{tab:cross_full}.} For each $\langle$model, curator$\rangle$ pair, the curated forget set $\ForgetData$ is constructed once and reused across SimNPO, UNDIAL, and RMU, so within each block only the unlearning algorithm changes. SimNPO rows repeat the corresponding entries of Table~\ref{tab:retrieval} (same runs), and the per-model cross evaluation for the EA curator is Table~\ref{tab:algos}. Table~\ref{tab:cross} reports the three-model averages.

\begin{table*}[!htp]
\centering
\footnotesize
\setlength{\tabcolsep}{4pt}
\renewcommand{\arraystretch}{1.1}
\caption{Per-model curator $\times$ unlearner cross evaluation at $|\ForgetReq|=50$. Metrics and columns follow Table~\ref{tab:retrieval}.}
\label{tab:cross_full}
\begin{tabular}{lll|cc|c|cccccc|c}
\toprule
& & & \multicolumn{2}{c|}{Net reduction (\%)} & QA & \multicolumn{7}{c}{Validation $\Delta$ (pp)} \\
\cmidrule(lr){4-5} \cmidrule(lr){7-13}
& & Model & F\,$\uparrow$ & R\,$\downarrow$ & $\Delta$ (pp) & GSM8K & BBH & WinoG & CoQA & HEval+ & LMBDA & \textit{Avg} \\
\midrule
\multirow{12}{*}{\rotatebox[origin=c]{90}{BM25 pre}}
& \multirow{4}{*}{\rotatebox[origin=c]{90}{\scriptsize SimNPO}}
  & \textsc{Llama-3.1-8B} & \cellcolor{cg50} 28.9\% & \cellcolor{cr75} 51.2\% & \cellcolor{cr25} -0.4 & \cellcolor{cr25} -0.1 & \cellcolor{cg25} +0.1 & \cellcolor{cr25} -0.6 & \cellcolor{cg50} +1.9 & \cellcolor{cg25} +1.2 & \cellcolor{cr25} -1.0 & \cellcolor{cg25} +0.3 \\
& & \textsc{Olmo-3-7B}    & \cellcolor{cg25} 1.0\% & \cellcolor{cr25} 2.9\% & \cellcolor{cr25} -0.1 & \cellcolor{cg50} +1.7 & \cellcolor{cr25} -0.1 & \cellcolor{cr25} -0.1 & \cellcolor{cr75} -3.8 & \cellcolor{cr25} -0.6 & \cellcolor{cneutral} -0.0 & \cellcolor{cr25} -0.5 \\
& & \textsc{Qwen3-8B}     & \cellcolor{cg25} 1.6\% & \cellcolor{cr25} 2.1\% & \cellcolor{cr75} -3.9 & \cellcolor{cr25} -1.4 & \cellcolor{cg25} +1.4 & \cellcolor{cg75} +3.2 & \cellcolor{cr75} -3.5 & \cellcolor{cr100} -6.7 & \cellcolor{cg75} +3.8 & \cellcolor{cr25} -0.5 \\
& & \textit{Average}      & \cellcolor{cavg} 10.5\% & \cellcolor{cavg} 18.7\% & \cellcolor{cavg} -1.5 & \cellcolor{cavg} +0.1 & \cellcolor{cavg} +0.5 & \cellcolor{cavg} +0.8 & \cellcolor{cavg} -1.8 & \cellcolor{cavg} -2.0 & \cellcolor{cavg} +0.9 & \cellcolor{cavg} -0.2 \\
\cmidrule(lr){2-13}
& \multirow{4}{*}{\rotatebox[origin=c]{90}{\scriptsize UNDIAL}}
  & \textsc{Llama-3.1-8B} & \cellcolor{cg100} 100.0\% & \cellcolor{cr100} 91.8\% & \cellcolor{cr25} -1.1 & \cellcolor{cg25} +0.8 & \cellcolor{cr25} -0.1 & \cellcolor{cr25} -0.6 & \cellcolor{cr100} -12.7 & \cellcolor{cg25} +0.6 & \cellcolor{cr50} -1.8 & \cellcolor{cr50} -2.3 \\
& & \textsc{Olmo-3-7B}    & \cellcolor{cg100} 97.8\% & \cellcolor{cr100} 79.6\% & \cellcolor{cr25} -0.1 & \cellcolor{cg50} +1.5 & \cellcolor{cr25} -0.1 & \cellcolor{cg25} +0.2 & \cellcolor{cr100} -5.1 & \cellcolor{cg75} +4.3 & \cellcolor{cr25} -1.3 & \cellcolor{cr25} -0.1 \\
& & \textsc{Qwen3-8B}     & \cellcolor{cg100} 96.7\% & \cellcolor{cr100} 79.0\% & \cellcolor{cr75} -3.5 & \cellcolor{cr25} -1.3 & \cellcolor{cr100} -5.1 & \cellcolor{cg75} +3.6 & \cellcolor{cr100} -5.3 & \cellcolor{cr25} -0.6 & \cellcolor{cg25} +0.3 & \cellcolor{cr25} -1.4 \\
& & \textit{Average}      & \cellcolor{cavg} 98.2\% & \cellcolor{cavg} 83.5\% & \cellcolor{cavg} -1.6 & \cellcolor{cavg} +0.3 & \cellcolor{cavg} -1.8 & \cellcolor{cavg} +1.1 & \cellcolor{cavg} -7.7 & \cellcolor{cavg} +1.4 & \cellcolor{cavg} -0.9 & \cellcolor{cavg} -1.3 \\
\cmidrule(lr){2-13}
& \multirow{4}{*}{\rotatebox[origin=c]{90}{\scriptsize RMU}}
  & \textsc{Llama-3.1-8B} & \cellcolor{cg100} 97.0\% & \cellcolor{cr100} 94.7\% & \cellcolor{cr100} -7.4 & \cellcolor{cr100} -48.7 & \cellcolor{cr100} -64.6 & \cellcolor{cr100} -6.8 & \cellcolor{cr100} -48.1 & \cellcolor{cr25} -0.6 & \cellcolor{cr100} -62.2 & \cellcolor{cr100} -38.5 \\
& & \textsc{Olmo-3-7B}    & \cellcolor{cg25} -0.2\% & \cellcolor{cr25} 6.9\% & \cellcolor{cg25} +0.3 & \cellcolor{cg25} +1.4 & \cellcolor{cneutral} -0.0 & \cellcolor{cg25} +1.4 & \cellcolor{cr25} -0.5 & \cellcolor{cg50} +3.0 & \cellcolor{cr25} -1.3 & \cellcolor{cg25} +0.7 \\
& & \textsc{Qwen3-8B}     & \cellcolor{cg25} 1.6\% & \cellcolor{cr25} 5.0\% & \cellcolor{cg25} +0.1 & \cellcolor{cr25} -1.1 & \cellcolor{cr25} -0.4 & \cellcolor{cr25} -1.2 & \cellcolor{cr25} -0.6 & \cellcolor{cg50} +3.0 & \cellcolor{cr25} -0.8 & \cellcolor{cr25} -0.2 \\
& & \textit{Average}      & \cellcolor{cavg} 32.8\% & \cellcolor{cavg} 35.5\% & \cellcolor{cavg} -2.3 & \cellcolor{cavg} -16.1 & \cellcolor{cavg} -21.7 & \cellcolor{cavg} -2.2 & \cellcolor{cavg} -16.4 & \cellcolor{cavg} +1.8 & \cellcolor{cavg} -21.4 & \cellcolor{cavg} -12.7 \\
\midrule
\multirow{12}{*}{\rotatebox[origin=c]{90}{BM25 mid}}
& \multirow{4}{*}{\rotatebox[origin=c]{90}{\scriptsize SimNPO}}
  & \textsc{Llama-3.1-8B} & \cellcolor{cg25} 16.7\% & \cellcolor{cr75} 54.0\% & \cellcolor{cr25} -1.1 & \cellcolor{cr25} -0.2 & \cellcolor{cr25} -0.4 & \cellcolor{cr25} -1.5 & \cellcolor{cr50} -2.4 & \cellcolor{cg50} +1.8 & \cellcolor{cr25} -1.4 & \cellcolor{cr25} -0.7 \\
& & \textsc{Olmo-3-7B}    & \cellcolor{cg25} 1.5\% & \cellcolor{cr25} 3.7\% & \cellcolor{cneutral} -0.0 & \cellcolor{cg25} +1.0 & \cellcolor{cr25} -0.2 & \cellcolor{cneutral} +0.0 & \cellcolor{cr50} -1.9 & \cellcolor{cr25} -1.2 & \cellcolor{cr25} -0.3 & \cellcolor{cr25} -0.4 \\
& & \textsc{Qwen3-8B}     & \cellcolor{cg25} 2.3\% & \cellcolor{cr25} 2.7\% & \cellcolor{cr100} -5.8 & \cellcolor{cr25} -1.1 & \cellcolor{cg25} +1.4 & \cellcolor{cg75} +3.6 & \cellcolor{cr25} -0.6 & \cellcolor{cr100} -10.4 & \cellcolor{cg75} +3.6 & \cellcolor{cr25} -0.6 \\
& & \textit{Average}      & \cellcolor{cavg} 6.8\% & \cellcolor{cavg} 20.1\% & \cellcolor{cavg} -2.3 & \cellcolor{cavg} -0.1 & \cellcolor{cavg} +0.3 & \cellcolor{cavg} +0.7 & \cellcolor{cavg} -1.6 & \cellcolor{cavg} -3.3 & \cellcolor{cavg} +0.6 & \cellcolor{cavg} -0.6 \\
\cmidrule(lr){2-13}
& \multirow{4}{*}{\rotatebox[origin=c]{90}{\scriptsize UNDIAL}}
  & \textsc{Llama-3.1-8B} & \cellcolor{cg100} 100.0\% & \cellcolor{cr100} 99.0\% & \cellcolor{cr50} -2.5 & \cellcolor{cr100} -5.8 & \cellcolor{cr75} -4.2 & \cellcolor{cr25} -0.3 & \cellcolor{cr100} -18.4 & \cellcolor{cr75} -4.3 & \cellcolor{cr75} -3.3 & \cellcolor{cr100} -6.1 \\
& & \textsc{Olmo-3-7B}    & \cellcolor{cg100} 86.9\% & \cellcolor{cr100} 80.3\% & \cellcolor{cr25} -1.0 & \cellcolor{cr25} -1.0 & \cellcolor{cr25} -0.9 & \cellcolor{cg25} +0.5 & \cellcolor{cr100} -12.3 & \cellcolor{cg50} +2.4 & \cellcolor{cr50} -1.8 & \cellcolor{cr50} -2.2 \\
& & \textsc{Qwen3-8B}     & \cellcolor{cg100} 91.9\% & \cellcolor{cr100} 80.1\% & \cellcolor{cr75} -4.1 & \cellcolor{cr50} -2.2 & \cellcolor{cr100} -27.6 & \cellcolor{cg50} +2.8 & \cellcolor{cr100} -14.2 & \cellcolor{cr50} -2.4 & \cellcolor{cr50} -1.7 & \cellcolor{cr100} -7.5 \\
& & \textit{Average}      & \cellcolor{cavg} 92.9\% & \cellcolor{cavg} 86.5\% & \cellcolor{cavg} -2.5 & \cellcolor{cavg} -3.0 & \cellcolor{cavg} -10.9 & \cellcolor{cavg} +1.0 & \cellcolor{cavg} -15.0 & \cellcolor{cavg} -1.4 & \cellcolor{cavg} -2.3 & \cellcolor{cavg} -5.3 \\
\cmidrule(lr){2-13}
& \multirow{4}{*}{\rotatebox[origin=c]{90}{\scriptsize RMU}}
  & \textsc{Llama-3.1-8B} & \cellcolor{cg100} 94.3\% & \cellcolor{cr100} 95.6\% & \cellcolor{cr100} -23.5 & \cellcolor{cr100} -48.8 & \cellcolor{cr100} -64.6 & \cellcolor{cr100} -17.4 & \cellcolor{cr100} -49.1 & \cellcolor{cr100} -24.4 & \cellcolor{cr100} -68.8 & \cellcolor{cr100} -45.5 \\
& & \textsc{Olmo-3-7B}    & \cellcolor{cg25} 12.9\% & \cellcolor{cr50} 39.8\% & \cellcolor{cr100} -6.4 & \cellcolor{cr75} -3.3 & \cellcolor{cr100} -26.7 & \cellcolor{cr25} -0.4 & \cellcolor{cr100} -20.7 & \cellcolor{cr25} -1.2 & \cellcolor{cr100} -35.4 & \cellcolor{cr100} -14.6 \\
& & \textsc{Qwen3-8B}     & \cellcolor{cg25} 5.0\% & \cellcolor{cr25} 9.3\% & \cellcolor{cr75} -4.0 & \cellcolor{cr50} -2.3 & \cellcolor{cr100} -5.1 & \cellcolor{cr50} -2.8 & \cellcolor{cr25} -0.1 & \cellcolor{cg100} +4.9 & \cellcolor{cr50} -2.9 & \cellcolor{cr25} -1.4 \\
& & \textit{Average}      & \cellcolor{cavg} 37.4\% & \cellcolor{cavg} 48.2\% & \cellcolor{cavg} -11.3 & \cellcolor{cavg} -18.1 & \cellcolor{cavg} -32.1 & \cellcolor{cavg} -6.9 & \cellcolor{cavg} -23.3 & \cellcolor{cavg} -6.9 & \cellcolor{cavg} -35.7 & \cellcolor{cavg} -20.5 \\
\midrule
\multirow{12}{*}{\rotatebox[origin=c]{90}{Infinigram mid}}
& \multirow{4}{*}{\rotatebox[origin=c]{90}{\scriptsize SimNPO}}
  & \textsc{Llama-3.1-8B} & \cellcolor{cg25} 19.4\% & \cellcolor{cr75} 55.7\% & \cellcolor{cr25} -0.8 & \cellcolor{cg25} +0.3 & \cellcolor{cr25} -0.4 & \cellcolor{cr25} -0.9 & \cellcolor{cg25} +0.9 & \cellcolor{cneutral} +0.0 & \cellcolor{cr25} -1.4 & \cellcolor{cr25} -0.2 \\
& & \textsc{Olmo-3-7B}    & \cellcolor{cg25} 1.6\% & \cellcolor{cr25} 3.2\% & \cellcolor{cr25} -0.1 & \cellcolor{cg25} +0.9 & \cellcolor{cr25} -0.3 & \cellcolor{cneutral} +0.0 & \cellcolor{cr25} -0.8 & \cellcolor{cr25} -0.6 & \cellcolor{cr25} -0.1 & \cellcolor{cr25} -0.2 \\
& & \textsc{Qwen3-8B}     & \cellcolor{cg25} 2.8\% & \cellcolor{cr25} 2.8\% & \cellcolor{cr100} -4.6 & \cellcolor{cr25} -1.0 & \cellcolor{cg25} +1.5 & \cellcolor{cg75} +3.9 & \cellcolor{cr75} -3.1 & \cellcolor{cr100} -9.8 & \cellcolor{cg75} +3.6 & \cellcolor{cr25} -0.8 \\
& & \textit{Average}      & \cellcolor{cavg} 7.9\% & \cellcolor{cavg} 20.6\% & \cellcolor{cavg} -1.8 & \cellcolor{cavg} +0.1 & \cellcolor{cavg} +0.3 & \cellcolor{cavg} +1.0 & \cellcolor{cavg} -1.0 & \cellcolor{cavg} -3.5 & \cellcolor{cavg} +0.7 & \cellcolor{cavg} -0.4 \\
\cmidrule(lr){2-13}
& \multirow{4}{*}{\rotatebox[origin=c]{90}{\scriptsize UNDIAL}}
  & \textsc{Llama-3.1-8B} & \cellcolor{cg100} 99.7\% & \cellcolor{cr100} 94.6\% & \cellcolor{cr50} -1.9 & \cellcolor{cg25} +1.0 & \cellcolor{cr25} -0.9 & \cellcolor{cg25} +0.2 & \cellcolor{cr100} -10.5 & \cellcolor{cg50} +3.0 & \cellcolor{cr50} -2.4 & \cellcolor{cr50} -1.6 \\
& & \textsc{Olmo-3-7B}    & \cellcolor{cg100} 75.8\% & \cellcolor{cr75} 69.4\% & \cellcolor{cg25} +0.1 & \cellcolor{cg25} +1.1 & \cellcolor{cr25} -0.8 & \cellcolor{cg25} +0.2 & \cellcolor{cr100} -8.3 & \cellcolor{cg50} +1.8 & \cellcolor{cr25} -1.4 & \cellcolor{cr25} -1.2 \\
& & \textsc{Qwen3-8B}     & \cellcolor{cg100} 78.0\% & \cellcolor{cr75} 72.9\% & \cellcolor{cr100} -4.5 & \cellcolor{cr25} -1.4 & \cellcolor{cr100} -10.5 & \cellcolor{cg50} +3.0 & \cellcolor{cr100} -6.7 & \cellcolor{cr25} -1.2 & \cellcolor{cr50} -2.4 & \cellcolor{cr75} -3.2 \\
& & \textit{Average}      & \cellcolor{cavg} 84.5\% & \cellcolor{cavg} 79.0\% & \cellcolor{cavg} -2.1 & \cellcolor{cavg} +0.2 & \cellcolor{cavg} -4.1 & \cellcolor{cavg} +1.1 & \cellcolor{cavg} -8.5 & \cellcolor{cavg} +1.2 & \cellcolor{cavg} -2.1 & \cellcolor{cavg} -2.0 \\
\cmidrule(lr){2-13}
& \multirow{4}{*}{\rotatebox[origin=c]{90}{\scriptsize RMU}}
  & \textsc{Llama-3.1-8B} & \cellcolor{cg100} 93.8\% & \cellcolor{cr100} 92.5\% & \cellcolor{cr100} -16.6 & \cellcolor{cr100} -49.6 & \cellcolor{cr100} -64.6 & \cellcolor{cr100} -10.8 & \cellcolor{cr100} -49.1 & \cellcolor{cr100} -6.7 & \cellcolor{cr100} -65.3 & \cellcolor{cr100} -41.0 \\
& & \textsc{Olmo-3-7B}    & \cellcolor{cg25} 6.5\% & \cellcolor{cr50} 32.5\% & \cellcolor{cr75} -3.1 & \cellcolor{cr25} -1.1 & \cellcolor{cr100} -17.7 & \cellcolor{cneutral} +0.0 & \cellcolor{cr100} -9.8 & \cellcolor{cr25} -0.6 & \cellcolor{cr100} -33.5 & \cellcolor{cr100} -10.5 \\
& & \textsc{Qwen3-8B}     & \cellcolor{cg25} 4.0\% & \cellcolor{cr25} 7.9\% & \cellcolor{cr50} -2.3 & \cellcolor{cr50} -2.1 & \cellcolor{cr75} -3.1 & \cellcolor{cr50} -2.7 & \cellcolor{cr25} -0.5 & \cellcolor{cg50} +3.0 & \cellcolor{cr50} -2.6 & \cellcolor{cr25} -1.3 \\
& & \textit{Average}      & \cellcolor{cavg} 34.8\% & \cellcolor{cavg} 44.3\% & \cellcolor{cavg} -7.3 & \cellcolor{cavg} -17.6 & \cellcolor{cavg} -28.5 & \cellcolor{cavg} -4.5 & \cellcolor{cavg} -19.8 & \cellcolor{cavg} -1.4 & \cellcolor{cavg} -33.8 & \cellcolor{cavg} -17.6 \\
\bottomrule
\end{tabular}
\end{table*}

\section{Curator Output Sizes}
\label{app:df_sizes}
The curators produce forget sets of different sizes; $|\ForgetData|$ is a property of the curator under test, and we keep the unlearner's optimization hyperparameters (such as epochs) fixed irrespective of $|\ForgetData|$. Table~\ref{tab:df_sizes} reports the exact number of prefix--suffix windows produced by each curator at $|\ForgetReq|=50$. The retain set passed to the unlearner is fixed at 1{,}646 WikiText rows for every configuration.

\begin{table}[!h]
\centering
\footnotesize
\caption{Number of curated forget-set windows at $|\ForgetReq|=50$.}
\label{tab:df_sizes}
\begin{tabular}{l|cccc}
\toprule
Model & BM25 pre & BM25 mid & Infini-gram mid & EA \\
\midrule
\textsc{Llama-3.1-8B} & 2{,}901 & 6{,}350 & 5{,}787 & 8{,}513 \\
\textsc{Olmo-3-7B}$^{\dagger}$ & 2{,}432 & 4{,}839 & 4{,}442 & 8{,}084 \\
\textsc{Nemotron-9B} & 2{,}835 & 6{,}324 & 6{,}011 & 8{,}840 \\
\textsc{Qwen3-8B} & 2{,}772 & 6{,}326 & 5{,}960 & 10{,}128 \\
\textsc{Gemma-3-12B} & 3{,}028 & 6{,}323 & 6{,}096 & 27{,}936 \\
\textsc{Olmo-3-32B} & 2{,}737 & 6{,}381 & 6{,}187 & 67{,}323 \\
\bottomrule
\end{tabular}

\smallskip
{\footnotesize $^{\dagger}$For \textsc{Olmo-3-7B}, only 38 works cleared the model-extractability threshold, so its forget pool contains 38 works rather than 50.}
\end{table}

\section{Songs vs.\ Books}
\label{app:domain}
Table~\ref{tab:domain} splits the Table~\ref{tab:retrieval} configurations by domain and averages over models. The qualitative result holds in both domains: retrieval curators do not produce strong selective suppression, and EA saturates target suppression without selectivity. Retain-side suppression under EA is higher for songs, which repeat internally and share choruses, idioms, and stock phrasing with other songs far more than books share text with other books. A 50-work request drawn from 4{,}616 songs and 50 books contains only one or two books on average, so the book column rests on few works per request.

\begin{table}[!h]
\centering
\footnotesize
\caption{Forget ($F$) and retain ($R$) net reduction (\%) by domain, SimNPO, $|\ForgetReq|=50$, averaged over models.}
\label{tab:domain}
\begin{tabular}{l|cc|cc}
\toprule
& \multicolumn{2}{c|}{Books} & \multicolumn{2}{c}{Songs} \\
Curator & $F\uparrow$ & $R\downarrow$ & $F\uparrow$ & $R\downarrow$ \\
\midrule
BM25 pre & 8.0 & 25.4 & 23.0 & 29.9 \\
BM25 mid & 18.6 & 31.3 & 19.5 & 26.7 \\
Infini-gram mid & 26.1 & 30.8 & 20.7 & 27.4 \\
EA & 99.1 & 66.9 & 99.6 & 99.5 \\
\bottomrule
\end{tabular}
\end{table}


\newpage

\end{document}